\documentclass[conference]{IEEEtran}
\IEEEoverridecommandlockouts
\usepackage{cite}
\usepackage{amsmath,amssymb,amsfonts}
\usepackage{graphicx, enumitem}
\usepackage{textcomp}
\usepackage[ruled,vlined,linesnumbered]{algorithm2e}
\usepackage[table,dvipsnames]{xcolor}
\usepackage{tikz}
\usepackage{bm}
\usepackage{makecell}
\usepackage{booktabs}
\usepackage{multirow}
\usepackage{url}
\usepackage{tikz}
\usepackage[none]{hyphenat}
\usepackage{caption}
\usepackage{subfig}
\usepackage{comment}
\usepackage{dblfloatfix}

\newcolumntype{L}[1]{>{\raggedright\let\newline\\\arraybackslash\hspace{0pt}}m{#1}}
\newcolumntype{C}[1]{>{\centering\let\newline\\\arraybackslash\hspace{0pt}}m{#1}}
\newcolumntype{R}[1]{>{\raggedleft\let\newline\\\arraybackslash\hspace{0pt}}m{#1}} 

\def\BibTeX{{\rm B\kern-.05em{\sc i\kern-.025em b}\kern-.08em
    T\kern-.1667em\lower.7ex\hbox{E}\kern-.125emX}}

\newcommand\blfootnote[1]{%
  \begingroup
  \renewcommand\thefootnote{}\footnote{#1}%
  \addtocounter{footnote}{-1}%
  \endgroup
}
\begin{document}

%Each paper should be between 6 and 8 pages (ideally 6 pages), including figures, tables and references. A maximum of two extra pages per paper is allowed (i.e, up to 10 pages), at an additional charge of US$100 per extra page. 
\title{Towards Improving Exploration in Self-Imitation Learning using Intrinsic Motivation}

\author{
\IEEEauthorblockN{Alain Andres\IEEEauthorrefmark{1}\IEEEauthorrefmark{2}\IEEEauthorrefmark{3},
	Esther Villar-Rodriguez\IEEEauthorrefmark{2}, and
	Javier Del Ser\IEEEauthorrefmark{2}\IEEEauthorrefmark{3}
	}
	\IEEEauthorblockA{\IEEEauthorrefmark{2}TECNALIA, Basque Research and Technology Alliance (BRTA), 48160 Derio, Bizkaia, Spain}
	\IEEEauthorblockA{\IEEEauthorrefmark{3}University of the Basque Country (UPV/EHU), 48013 Bilbao, Bizkaia, Spain}
	\IEEEauthorblockA{\IEEEauthorrefmark{1}Corresponding author; email: alain.andres@tecnalia.com}}
\maketitle

\begin{abstract}
Reinforcement Learning has emerged as a strong alternative to solve optimization tasks efficiently. The use of these algorithms highly depends on the feedback signals provided by the environment in charge of informing about how good (or bad) the decisions made by the learned agent are. Unfortunately, in a broad range of problems the design of a good reward function is not trivial, so in such cases sparse reward signals are instead adopted. The lack of a dense reward function poses new challenges, mostly related to exploration. Imitation Learning has addressed those problems by leveraging demonstrations from experts. In the absence of an expert (and its subsequent demonstrations), an option is to prioritize well-suited exploration experiences collected by the agent in order to bootstrap its learning process with good exploration behaviors. However, this solution highly depends on the ability of the agent to discover such trajectories in the early stages of its learning process. To tackle this issue, we propose to combine imitation learning with intrinsic motivation, two of the most widely adopted techniques to address problems with sparse reward. In this work intrinsic motivation is used to encourage the agent to explore the environment based on its curiosity, whereas imitation learning allows repeating the most promising experiences to accelerate the learning process. This combination is shown to yield an improved performance and better generalization in procedurally-generated environments, outperforming previously reported self-imitation learning methods and achieving equal or better sample efficiency with respect to intrinsic motivation in isolation.\blfootnote{Work published at the IEEE Symposium on Adaptive Dynamic Programming and Reinforcement Learning (IEEE ADPRL), 2022.}

\end{abstract}

\begin{IEEEkeywords}
Reinforcement Learning, Intrinsic Motivation, Self Imitation Learning, Sparse Rewards, Generalization
\end{IEEEkeywords}

\section{Introduction} \label{sec:intro}

Reinforcement Learning has captured the attention of the research community due to the manifold applications in which this kind of learning problems can be formulated, including games \cite{mnih2015human}, healthcare \cite{gottesman2019guidelines}, industry \cite{nian2020review} and robotics \cite{kober2013reinforcement}, among others. This momentum has flourished from studies reporting the proven superiority of Deep Reinforcement Learning (DRL) over humans in certain tasks \cite{silver2017mastering,vinyals2019grandmaster}, and its potential to train generalist agents \cite{reed2022generalist}. 

Unfortunately, the large number of samples usually required for DRL makes its application infeasible in some real-world settings. Consequently, Imitation Learning (IL) and Transfer Learning strategies have been proposed over the years to accelerate the learning process and decrease the required amount of training data \cite{hua2021learning,nair2020awac}. The same strategy of using expert demonstrations has been used to address exploration issues in problems with sparse rewards and help the agent learn without any feedback signal from the environment \cite{salimans2018learning}. Nevertheless, such expert demonstrations are not always available in practice. This motivated the idea of the agent interacting with the environment and storing experiences entailing good exploration properties, namely, \textit{self-Imitation Learning} (self-IL \footnote{For the sake of clarity, in the paper we refer to this kind of methods as \emph{self-IL}, whereas we use SIL to refer to the approach presented in \cite{oh2018self}.}). Although it effectively alleviates the need for expert demonstrations, self-imitation learning methods have a high sensitivity to the discovery of sufficiently good trajectories in first instance which can be difficult to obtain depending on the scenario. On the other hand, exploration issues have been also tackled from the perspective of Intrinsic Motivation (IM), in which the agent is encouraged to interact with the environment based on its inner curiosity rather than on externally signalled information about the quality of the interaction with respect to the formulated goal. %However, this does not solve the problem of establishing a proper balance between exploration and exploitation during the learning process \cite{andres2022collaborative}.

% PROPOSAL AND HOW DIFFERENT IT IS
In the absence of expert demonstrations and with the aim of maximizing the sample-efficiency in sparse rewards problems, this work proposes a framework that combines both self-IL and IM towards achieving a better management of the agent's exploration capabilities in sparse problems. Both strategies are commonly used to address tasks demanding high exploration, yet in a separate fashion. Our research hypothesis is that these strategies can be synergistic: IM can be used to generate intrinsic rewards and foster exploration directly, so that the limitation of discovering diverse and valuable trajectories exposed by self-IL can be alleviated. At the same time, self-IL can be effectively used to replay and prioritize good experiences in the agent's learning process. We envision that, by combining both approaches, we simultaneously boost generalizable knowledge by reinforcing those experiences attractive in term of objective as well as those novel potentially leading to better outcomes.

%y combining both approaches, we enhance exploration reinforcing those experiences that are more attractive in terms of the objective which might be difficult to reproduce just relying into IM due to curiosity vanishing and the emergent disentanglement/detachment \cite{ecoffet2021first}.

%the exploration is not only fenced in a given episode that might be difficult to reproduce again with just IM, due to curiosity vanishing and the disentanglement/detachment to which IM mechanisms are subject \cite{ecoffet2021first}. 

This is not the first time that both strategies have been mixed to see whether they are complementary \cite{oh2018self,sovrano2019combining,ning2021co}. Previous research studies were carried out in singleton environments \cite{oh2018self,sovrano2019combining} with different kinds of prioritization strategies. 
Contrarily, procedurally-generated environments are created depending on specified attributes (or seeds) that govern the characteristics/dynamics of the generated environment (starting point, configuration of the objects, distance to the goal, etc) to measure the sample-efficiency and generalization capabilities. That is why, recently, hybrid IL-IM algorithms were evaluated over these environments yet being analyzed in low to moderate difficulty tasks, leaving the open question whether such algorithms can solve harder exploration tasks \cite{ning2021co}. Besides, in the absence of an ablation study it is difficult to determine the contribution of each element of the approach to expose their limitations. In this work we evaluate our framework in procedurally-generated environments from MiniGrid \cite{gym_minigrid}, where state-of-the-art algorithms and even IM-based solutions have difficulties to learn \cite{aa2022cdmake}. We build our approach on top of a novel self-IL approach called RAPID \cite{zha2021rank}, which retains good exploration behaviors by inspecting the entire episode, and takes into account both extrinsic and intrinsic feedback signals for replay. We evaluate the performance of our proposal in procedurally-generated hard-exploration environments, beating previously reported state-of-the-art self-IL methods results, and also achieving an equal or better sample efficiency with respect to IM approaches in isolation. In addition, we identify several weaknesses that might well hinder a better performance and spur future research directions.

%We evaluate the performance of our proposal in procedurally-generated hard-exploration environments where the agent has to learn not only to solve a given task, but also to generalize their knowledge to solve newly produced task instances. Our results reveal that RAPID and SIL \cite{oh2018self} in isolation do not suffice to learn and solve problems with very sparse reward. By contrast, our proposed approach manages to get an optimal policy even when facing discouraging contexts later exposed in the discussion. Such identified weaknesses spur several research directions to be pursued in the future, which will be also discussed in the closing part of the manuscript.

% Sections description
The rest of this manuscript is structured as follows: Section \ref{sec:back} presents the background related to IL, self-IL and IM. Next, Section \ref{sec:IM_IL} describes how self-IL and IM are combined together to foster a more efficient and synergistic exploration. Section \ref{sec:exp} follows by detailing the experimental setup. Section \ref{sec:results} examines the reported results and discusses the limitations. Finally, Section \ref{sec:conc} draws conclusions and future research work to be developed in this research line.

\section{Related Work} \label{sec:back}

In this section we briefly review the literature related to IL and self-IL (Section \ref{sec:back}.a), IM (Section \ref{sec:back}.b) and IL+IM approaches (Section \ref{sec:back}.c) to deal with sparse reward problems from a sample-efficient perspective. %Section \ref{sec:back}.d ends by posing the contribution of this work framed within the literature review introduced previously.

\paragraph{Imitation Learning (IL)} The experience replay buffers have been employed to stabilize the learning process when using neural network in Reinforcement Learning \cite{mnih2015human}. Instead of just focusing on the whole history of experiences, different works have proposed to prioritize those samples based on the TD-error \cite{schaul2015prioritized} or high extrinsic trajectories \cite{zha2021simplifying}. The goal is to sample the most promising experiences more frequently, potentially expediting and enhancing the overall knowledge learned by the agent. Following the idea of improving the sample efficiency, IL methods have resorted to expert demonstrations to accelerate the learning stage process, by forcing the agent to learn the inherent decision making hidden in those examples \cite{hester2018deep, vecerik2017leveraging}. However, collecting expert demonstrations is not easy to achieve and most of the time the agent is fed with suboptimal demonstrations that hinder the learning of an optimal policy \cite{nair2018overcoming,ning2020reinforcement}. When lacking an expert capable of providing examples, self-IL was proposed so that an agent, without any kind of prior knowledge, is responsible for collecting and imitating good experiences improvised in the past \cite{abolafia2018neural,oh2018self,chen2020self}. Interestingly, self-IL approaches aim to deal with exploration-exploitation dilemma by encouraging the agent to exploit the information that has not been previously learned, so as to achieve a better exploration strategy and ultimately, a near-optimal performance. Despite their proven efficiency at hard exploration problems \cite{oh2018self,guo2020memory,ecoffet2021first}, these methods tend to struggle in tasks characterized by very sparse rewards due to their reduced capability to find good trajectories early in the learning process by just naive exploration \cite{pshikhachev2022self}. This can be the reason why self-IL solutions have so far been evaluated mainly over non-procedurally-generated environments, where the generalization capabilities of the agent are not tested since the observation's features do not change from level to level \cite{guo2020memory,ecoffet2021first}. Alternatively, new approaches like RAPID \cite{zha2021rank} have emerged to deal with procedurally-generated environments by ranking the episodes not only in terms of their extrinsic rewards, but also by considering exploration scores related to the level. As a result of this multi-criteria ranking, the agent can imitate those experiences and overcome the exploration needs of very sparse tasks, as opposed to other approaches that prioritize episodes based just on their achieved extrinsic reward \cite{oh2018self}.

\paragraph{Intrinsic Motivation (IM)} Another research direction to enhance the exploration of the agent and thereby achieve a good overall performance for a given task is to use IM mechanisms. Some of the IM methods were not originally designed to tackle procedurally-generated environments, such as Counts \cite{bellemare2016unifying}, Random Network Distillation (RND \cite{burda2018exploration}), or Intrinsic Curiosity Module (ICM \cite{pathak2017curiosity}). Lately new approaches have emerged with the aim to learn more robust and generalized knowledge: this is the case of BeBold/NoveId \cite{zhang2021noveld}, MADE \cite{zhang2021made}, RIDE \cite{raileanu2020ride} or AGAC \cite{flet2021adversarially}. One of the common aspects to all these methods is that the generated rewards are commonly used with on-policy algorithms (i.e., A2C/A3C, PPO, IMPALA), which discard the collected experiences after their respective optimization updates. This hinders a good sample efficiency, even more so when considering disentanglement/derailment \cite{ecoffet2021first} and catastrophic forgetting due to the fades of intrinsic rewards \cite{huang2022solving}.

\paragraph{Combined IL+IM approaches} The idea of combining IL and IM strategies is not new. Previous works have analyzed if they are complementary to each other. However, most of them have been studied at singleton environments where the generalization capabilities have not been assessed. SIL \cite{oh2018self} used a count-based intrinsic reward to augment the exploration capabilities in singleton Apple-Gold mazes. Similarly, DTSIL \cite{guo2020memory} employed the same environment and evaluated its generalization ability by training over 12 random mazes and by evaluating the trained agent over 6 test instances. However, it requires a hierarchical policy where they do not use any kind of IM technique: they only combine DTSIL and IM to evaluate the exploration improvements over some Atari's (non procedurally-generated) environments. The approach presented in \cite{sovrano2019combining} was tested in Atari's Montezuma, Solaris and Venture environments, in which different prioritizing strategies were studied with prioritized experience replay and IM methods. Recently, \cite{ning2021co} evaluated its solution with SIL and BeBold in relatively easy sparse MiniGrid environments. Nevertheless, neither \cite{ning2021co} nor \cite{zha2021rank} address hard exploration environments as in other works related to IM \cite{zhang2021noveld, flet2021adversarially}. At this point, we note that by \emph{hard exploration} we refer to different configurations of the environments that make them more difficult to be solved. For instance, using more rooms in Multiroom; larger room size in KeyCorridor; or facing ObstructedMaze. In these tasks the generalization and learning requirements are of utmost importance and call for better exploration strategies during the agent's learning process.

\paragraph{Contribution} We formally propose the use of a self-IL strategy together with IM, showing that this combination succeeds at solving hard exploration environments. Our proposal builds upon RAPID \cite{zha2021rank}, in which we combine its ranking strategy for past stored episodes with intrinsic motivation mechanisms. This manuscript steps beyond the state of the art by showing that previous approaches did not consider \cite{ning2021co} or failed to solve \cite{zha2021rank} procedurally-generated environments with hard exploration requirements (see Section \ref{sec:exp}). Our experiments in scenarios with different learning challenges validate this claim and prove that the combination of RAPID and IM methods can meet the exploration requirements needed to solve hard exploration tasks.
%Our experiments in scenarios with different learning challenges inform with evidence this claim, and prove that the combination of RAPID and IM methods can meet the exploration requirements needed to solve hard exploration tasks.

%PLR \cite{jiang2021prioritizedlevelreplay} proposes a framework to selectively sample next training levels by prioritizing those with higher potential estimated learning. Focused on procedurally-generated environments for generalization by curriculum increasing difficulty levels. The learning potential is calculated with the average magnitude of GAE over a given trajectory in an episode/level. 
%It is being also a lot of interest in the field of self-supervision learning \cite{laskin2020curl}, where IM is also being used to improve the sample-efficiency and also the generalization capabilities when being used as an auxiliary objective \cite{zhao2021intrinsically,wu2022generalizing}.

\section{Synergistic exploration with Self Imitation Learning and Intrinsic Motivation through Ranked Episodes}\label{sec:IM_IL}

Self-IL methods allow the agent itself to reproduce self-collected past experiences to induce a better exploration behavior in its learning process \cite{oh2018self}. The core idea is to replay experiences that potentially improve the performance of the agent, even though that information was not persisted in the agent because experiences were not exploited enough. This makes the agent explore more effectively. This behavior is emphasized with on-policy algorithms, where samples are discarded after the optimization step. However, self-IL methods are highly sensitive to the discovery of good samples, and is subject to the capability of the algorithms to find them on their own (e.g., by establishing stochastically sampled policies). This might not be sufficient in very sparse reward problems, where the probability of achieving a non-zero rewarded episode is substantially low. This is the reason why IM, combined with self-IL, can imprint the explorative behavior needed to collect good episodes based on the agent's curiosity rather than on the extrinsic feedback signal.

In this section we explain a self-IL and a IM approach, namely, RAPID (Section \ref{sec:RAPID}) and BeBold (Section \ref{subsec:bebold}), which lie at the heart of the proposed framework described in Section \ref{sec:proposed}.

%\subsection{RAnking the ePIsoDes (RAPID)} \label{sec:RAPID}
\subsection{Ranking the episodes (RAPID)} \label{sec:RAPID}

RAPID \cite{zha2021rank} was proposed to endow an agent with a more general criterion to detect good exploration behaviors. To this end, RAPID assigns episodic exploration scores to each episode's experiences instead of relying only on state-level intrinsic rewards. The overall episodic score is calculated by a weighted sum of three scores:
\begin{equation} \label{eq:rapid}
    S= w_0S_{ext} + w_1S_{local} + w_2S_{global},
\end{equation}
where $S_{ext}$ is the total extrinsic reward of the episode, $S_{local}$ encourages the exploration inside the episode by maximizing the diversity across visited states, and $S_{global}$ models a long-term exploration view by using the average curiosity of the states inside the episode from a visitation count that takes not only the current episode but the whole training process into account. Based on these scores, the most promising episodes are kept in the replay buffer, so that highly ranked episodes are replayed and ultimately imitated by the agent to enhance its exploration. 

%\subsection{BEyond the BOundary of expLoreD regions (BeBold)} \label{subsec:bebold}
\subsection{Beyond the boundary of explored regions (BeBold)} \label{subsec:bebold}

BeBold \cite{zhang2021noveld} proposes an intrinsic motivation method that circumvent the \textit{short-sightedness} and \textit{detachment} problems by issuing a reward signal whenever the next state $s_{t+1}$ is \emph{more novel} than the current one $s_t$:
\begin{equation} \label{eq:bebold}
r^i_t = \max\left(\frac{1}{N(s_{t+1})}-\frac{1}{N(s_t)},0\right),
\end{equation}
where $N(s_t)$ denotes the number of times the agent has visited state $s_t$ during training. Moreover, BeBold minimizes the undesired behavior of the agent going back and forth by imposing that a state receives a reward only once per episode.

\subsection{Proposed Framework} \label{sec:proposed}

As stated in \cite{levine2022understanding}, IM-based exploration methods provide an auxiliary objective to collect more diverse data rather than learning to utilize it. Our proposed framework aims to exploit efficiently those collected diverse data during the agent's interaction (used in the on-policy RL updates) and go one step further by strengthen its knowledge with the off-policy/supervised loss, which replays and prioritizes the most promising episodes from which to learn.
%First of all, we generate intrinsic rewards to give more guarantees and increase the probability of discovering good episodes. After that, we persist such episodic information in a replay buffer so that the collected experiences are used to improve the agent more than once. Besides, we rank the episodes inside the buffer with not only extrinsic feedback from the environment, but also by other measures that enhance the episodes with explorative attractiveness, so that what the replay buffer also fosters exploration behavior. 

Bearing this in mind, our framework first generates an intrinsic reward $r^i$ as in Expression \eqref{eq:bebold} at each step, which is weighted by a factor $\beta$ before being added to the extrinsic reward $r^e$ given by the environment, yielding the overall reward $r_t=r_t^e + \beta r_t^i$. This reward is used by the selected RL algorithm to maximize the discounted return, $G_t = \sum_{k=t}^\infty \gamma^{k-t}r_{k}$. Then, the most promising experiences $(s,a)$ are stored in a buffer of limited size. The criterion to keep episodes is driven by (1) an extrinsic component such as the non-biased Monte Carlo extrinsic return of the episode, $G = \sum_{t=0}^T r_t^e$ and also by (2) scores that augment the exploration behavior in the episode, i.e., local and global scores exposed by RAPID as per Expression \eqref{eq:rapid}. After a given number of steps, a batch of experiences is sampled from the buffer and the policy is enforced to match previously executed actions by \emph{behavioral cloning}\footnote{Behavioral cloning is selected for its relative simplicity \cite{osa2018algorithmic}, but other alternatives for policy matching can be equally considered.}. 

With this proposal, the RL update will be strengthened with a intrinsic reward that promotes exploration and augments the probability of sampling good episodes for the ranking buffer, while the latter stores and persists previous highly-ranked episodes to keep improving the agent even when the on-policy updates are not enough and when its intrinsic exploration bonus vanishes. Likewise, both losses will foster exploration while maximizing the exploitation of the given task.

\section{Experimental Setup}\label{sec:exp}

This section describes in depth the environments considered to evaluate the results of the proposed framework (Section \ref{subsec:exp_env}), the baselines for comparison with their selected hyperparameters and neural network architectures (Section \ref{subsec:baselines_hyperparams}), and the adopted evaluation methodology (Section \ref{subsec:eval_code}). In order to guarantee transparency and reproducibility of the experiments later discussed, software scripts and configuration files have been made available in a public GitHub repository: \url{https://github.com/aklein1995/exploration_sil_im}.

\subsection{Environments} \label{subsec:exp_env}

Performance evaluations are carried out over procedurally-generated environments, where the agent position and the configuration of objects can randomly change depending on a numerical seed. The goal is to learn a policy capable of solving unseen instances of such environments. In the present work we employ some of the hard-exploration scenarios provided in MiniGrid \cite{gym_minigrid}. Environments in this benchmark are built on a grid with a discrete state/observation space. Observations are egocentric and partially observable to the agent in the form of $7\times 7$ tiles. Moreover, each object is represented with a compact encoding of 3 values. Thus, an observation is a value tensor of size $7\times 7\times 3$. Moreover, 7 different actions are available to solve any given scenario: \texttt{turn left}, \texttt{turn right}, \texttt{move forward}, \texttt{pick up} (an object, e.g. keys or balls), \texttt{drop} the object (if carried), \texttt{toggle} (open doors, interact with objects) and \texttt{done}. Specifically, we evaluate the framework over the following scenarios (for further information about the environments and their tasks, please refer to \cite{gym_minigrid}):
\begin{itemize}
    \item MultiRoom (\texttt{MN7S8} and \texttt{MN12S10}).
    \item KeyCorridor (\texttt{KS4R3}).
    \item ObstructedMaze (\texttt{O2Dlh}).
\end{itemize}

The criteria to select these environments relies on their difficulty as exposed in \cite{zha2021rank}. In that work \texttt{MN12S10} and \texttt{KS4R3} were identified as the most difficult analyzed scenarios: the first was solved by RAPID and RIDE, while the latter remained unsolved for the given train steps by any of the baselines under consideration. In the case of \cite{ning2021co} where the performance of SIL+BeBold was analyzed, the most difficult environments were \texttt{KS3R3} and \texttt{MN6S}, which are more easily solvable than \texttt{KS4R3} and \texttt{MN12S10}. Additionally, we include another very hard exploration scenario, not considered in the aforementioned works, which possesses different characteristics and requirements than the previous environments: \texttt{O2Dlh}.

\subsection{Baselines and Hyperparameters} \label{subsec:baselines_hyperparams}

We select RAPID \cite{zha2021rank} and SIL \cite{oh2018self} as self-IL methods. The intrinsic reward function proposed by BeBold \cite{zhang2021noveld} is chosen, in which novelty is calculated by using visitation counts $N(s)$. All strategies use PPO as RL algorithm, which uses a number of steps equal to $128$ and $4$ minibatches of size $32$ for training (one unique agent). Each train step comprises $4$ epochs, where optimization updates are carried out with a clipping factor of $0.2$, a learning rate of $10^{-4}$, $\gamma=0.99$ and $\lambda=0.95$ for the advantages calculation with GAE. Furthermore, the loss function is weighted with a entropy coefficient of $0.01$ and a value coefficient of $0.5$. We employ 2 independent fully-connected layers for the actor and the critic, each with $64$ neurons.

By default, we configure RAPID as in its original paper, with a buffer size of $10^4$ experiences, batch size of $256$, $5$ off-policy updates after each episode completion. Weights to rank the replay buffer experiences -- Expression \eqref{eq:rapid} -- are set to $w_0=1$, $w_1=0.1$ and $w_2=0.001$. In the case of SIL, to be as fair as possible with respect to RAPID, we use the same replay buffer size ($10^4$) and the same off-policy update ratio ($5$). Moreover, we establish a loss weight of $0.1$ and a value loss weight of $0.01$. Regarding PER \cite{schaul2015prioritized}, we select a prioritization exponent $\alpha=0.6$ and a bias correction factor $\beta=0.1$. All these parameter values were chosen according to the supplementary material provided in \cite{oh2018self}, and taking into account that we target at solving hard exploration environments. On the other hand, the intrinsic reward when using BeBold is computed as described in Section \ref{subsec:bebold}, using an intrinsic coefficient of $0.005$. These values were tailored based on the results of a grid search carried out over scenario \texttt{MN7S8}, which can be found in the GitHub repository.

\subsection{Evaluation Methodology} \label{subsec:eval_code}

We report the mean and standard deviation of the average return computed over the past 100 episodes for each experiment, performing 3 different runs (with different seeds) to account for the statistical variability of the results. 

\section{Results \& Discussion} \label{sec:results}

This section is devoted to present insights about how \textit{intrinsic motivation} can help \textit{self-imitation learning} methods to boost their performance (Section \ref{subsec:results}) and discuss potential limitations (Section \ref{subsec:limitations}).

\subsection{Results}\label{subsec:results}
%\vspace{2mm}

\begin{figure}[h]
    \centering
    \subfloat{\includegraphics[width=1.0\columnwidth]{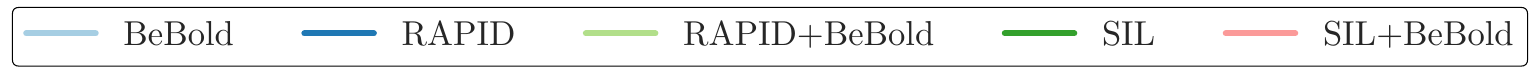}}
    \\
    \subfloat{\includegraphics[width=0.5\columnwidth]{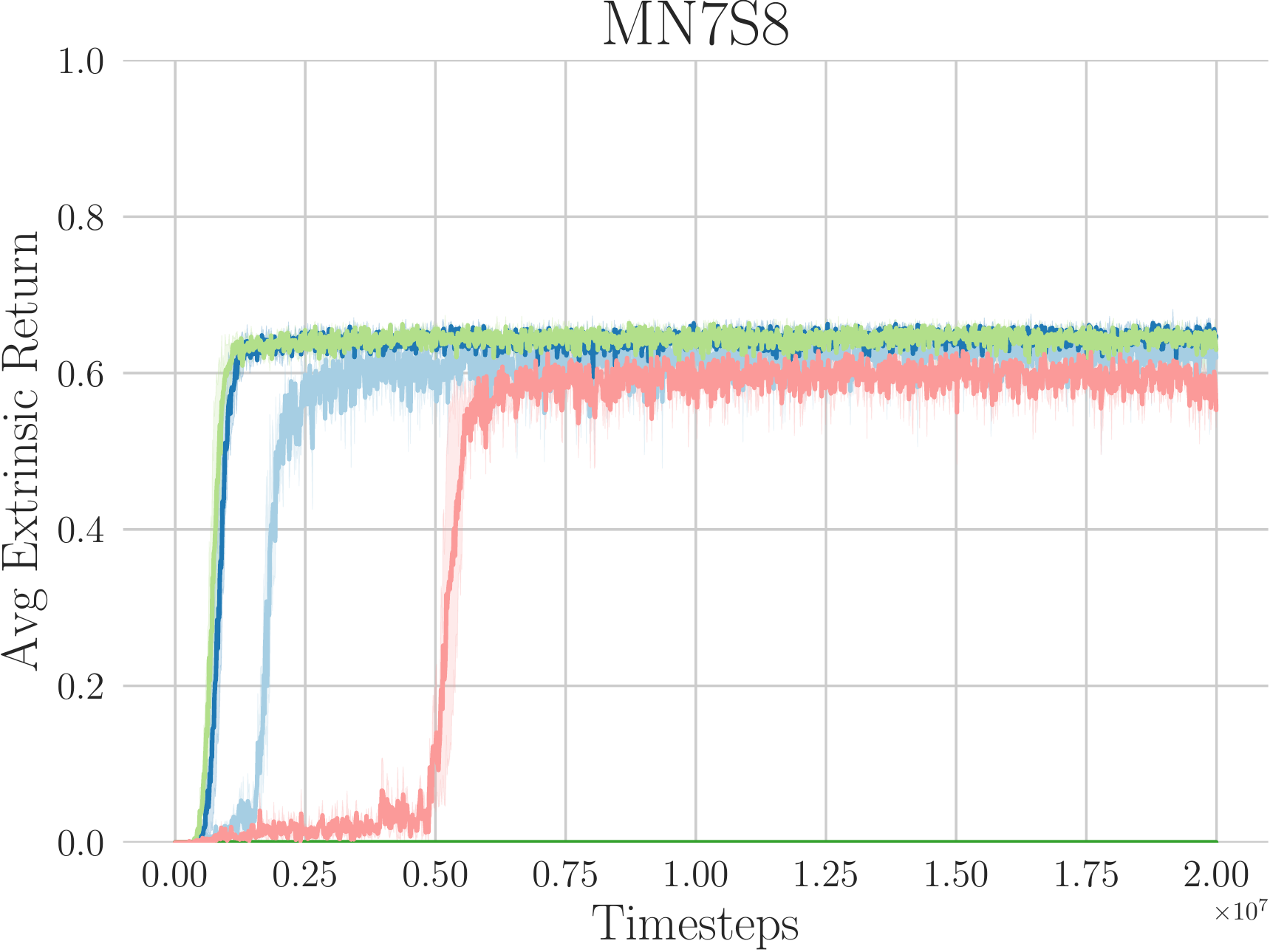}} 
    \subfloat{\includegraphics[width=0.5\columnwidth]{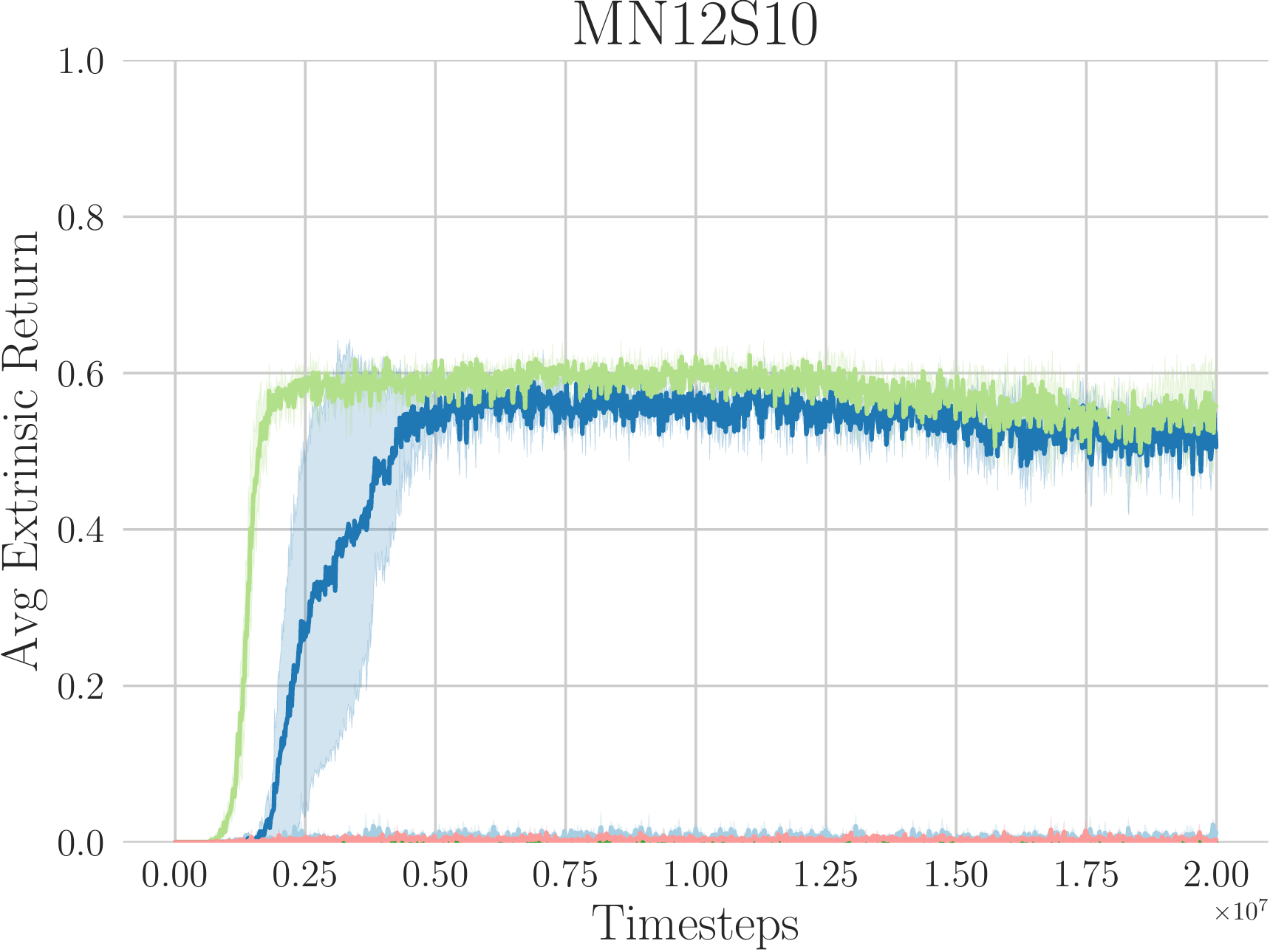}}
    \\
    \subfloat{\includegraphics[width=0.5\columnwidth]{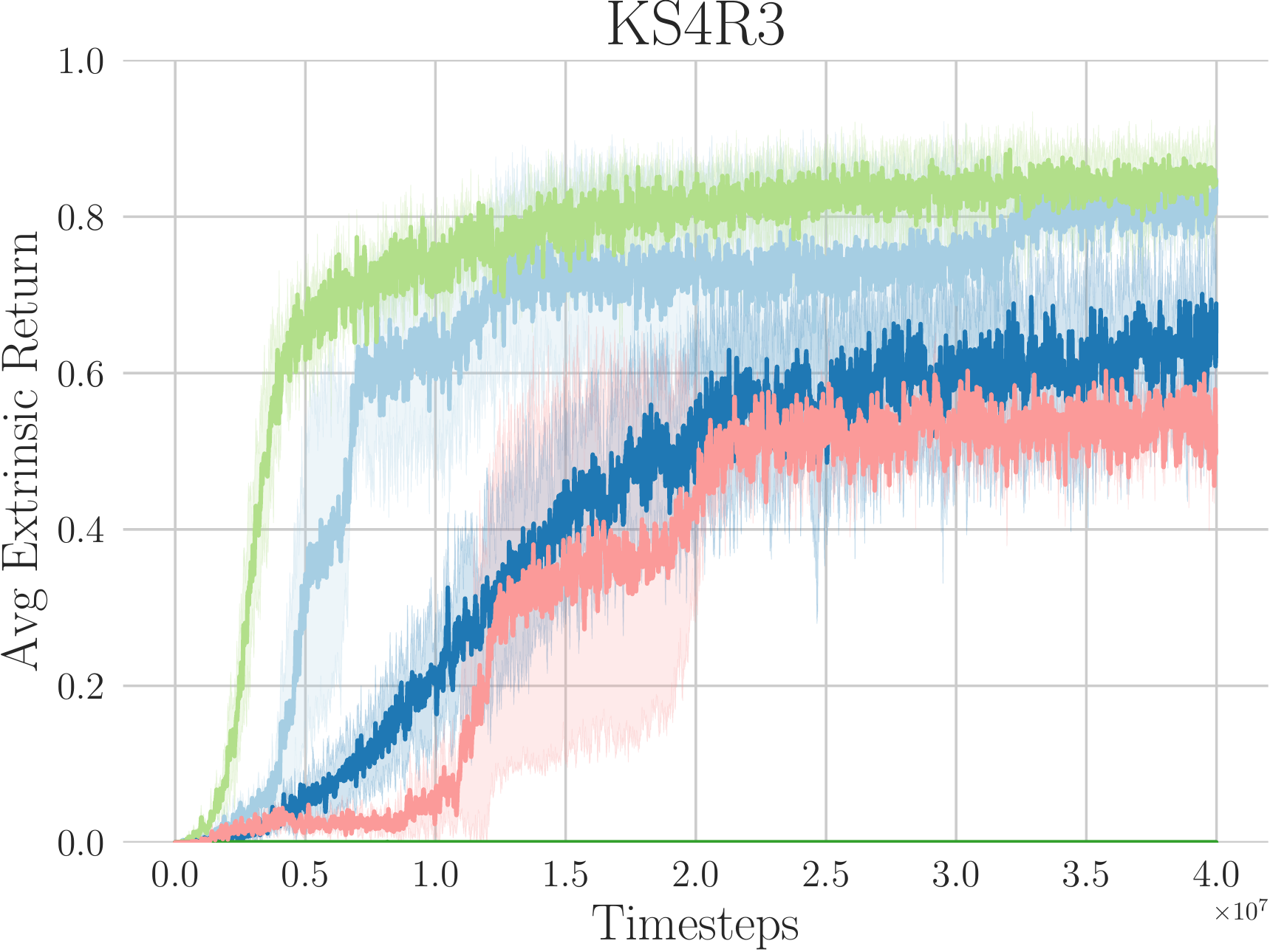}} 
    \subfloat{\includegraphics[width=0.5\columnwidth]{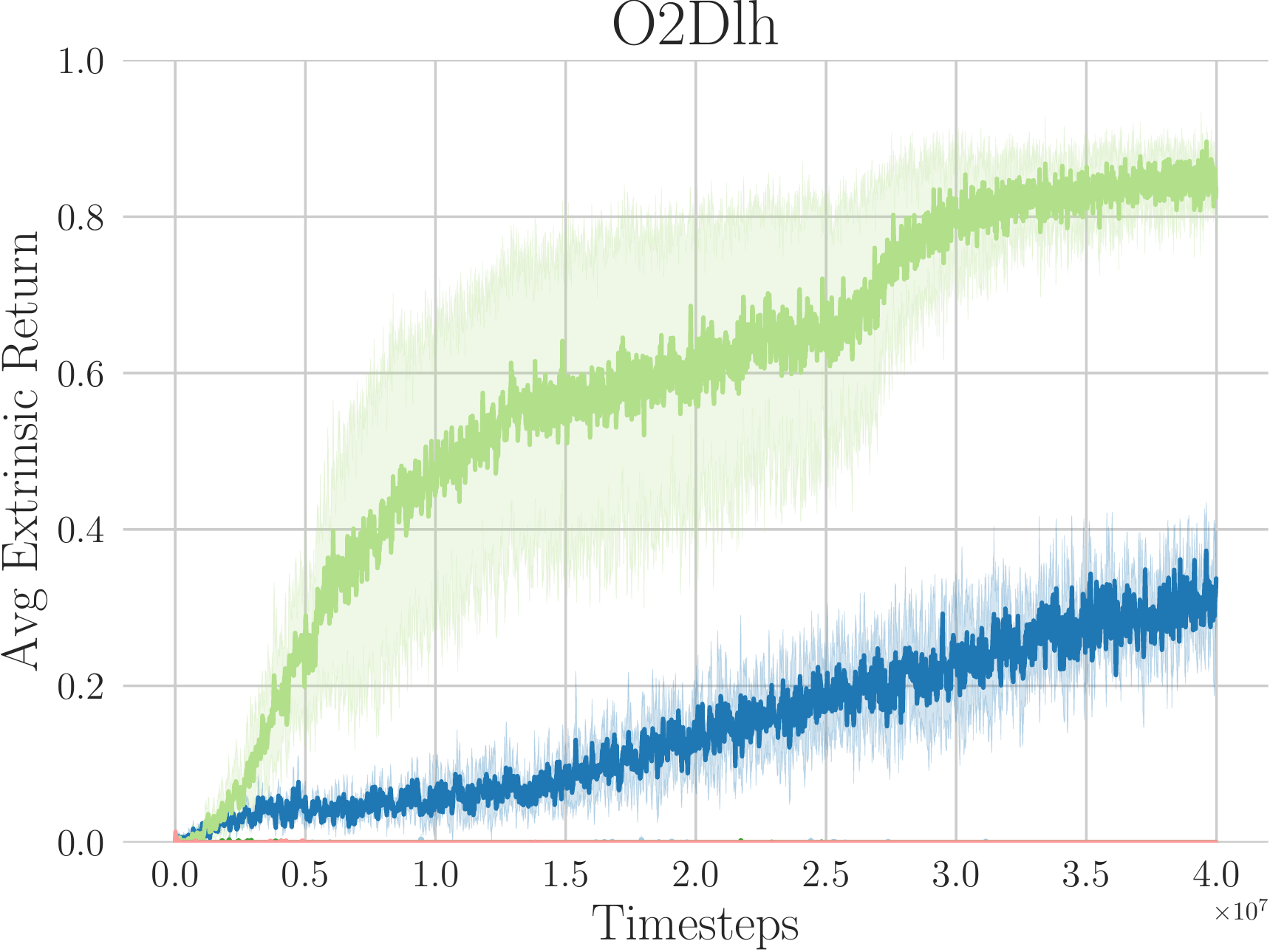}}
    \caption{Results over multiple procedurally-generated hard exploration environments in MiniGrid. Both RAPID and SIL always achieve better results when combined with BeBold.}
    \label{fig:core_results}
\end{figure}

\textbf{Analyzing the performance of self-IL and IM techniques independently and when being combined:} to begin with, Figure \ref{fig:core_results} analyzes the actual impact on the performance of the agent when using IM and self-IL techniques either independently or in combination. We observe that BeBold shows a good behavior only in 2 out of the 4 environments under consideration (namely, \texttt{MN7S8} and \texttt{KS4R3}). However, it completely fails when dealing with the challenging scenarios of MultiRoom and ObstructedMaze series (corr. \texttt{MN12S10} and \texttt{O2Dlh}). When using just SIL, it performs poorly in all scenarios\footnote{As mentioned in Section \ref{sec:back} and Section \ref{subsec:exp_env}, other works have analyzed how complementary SIL and some IM techniques were, but at sparse reward problems that are not so hard as the ones herein considered \cite{ning2021co}.}. 

With reference to RAPID, it is capable of solving \texttt{MultiRoom} environments (as expected), but struggles over \texttt{KS4R3} and \texttt{O2Dlh}, which are assumed to have larger state spaces and an increasing difficulty from the perspective of exploration. On top of the self-IL approaches, BeBold fosters the exploration and consequently renders some valuable learning when using SIL. However, results are worse than using BeBold in isolation, which suggests that the SIL prioritization mechanisms are not working properly. Contrarily, results are outstanding when combined with RAPID, reducing drastically the number of samples to achieve the same performance and attaining a better overall learning when compared to using RAPID in its na\"ive version. Besides these improvements, it is interesting to notice that the benefits of using IM remain even when the latter is not enough to learn in isolation: BeBold does not capture any knowledge over \texttt{MN12S10} and \texttt{O2Dlh}, but it augments the capabilities of RAPID when used in those scenarios.

% 2. Speak about different Intrinsic Motivations and how they impact the learning
\textbf{Evaluation of RAPID with other IM strategies:} A key aspect of this work is the capacity of IM to enhance the agent's exploration while learning. Therefore, it is of utmost importance to assess the sensitivity of the proposed self-IL+IM combination with respect to the selection of the IM approach. With that in mind, and considering that the current implementation is based on BeBold's tabular version (see Section \ref{subsec:bebold}), we now evaluate the agent's performance with other two visitation counts strategies: \emph{counts} (i.e. $r_i=1/\sqrt{N(s')}$) and \emph{counts1st}, which is the same as \emph{counts} but with episodic restriction. This second set of experiments allows comparing very similar IM strategies that have proven to have different results due to their intrinsic reward generation schema \cite{aa2022cdmake,zhang2021noveld}. 

\begin{figure}[h]
    \centering
    \subfloat{\includegraphics[width=1.0\columnwidth]{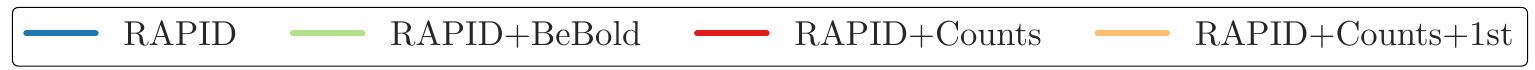}}
    \\
    \subfloat{\includegraphics[width=0.5\columnwidth]{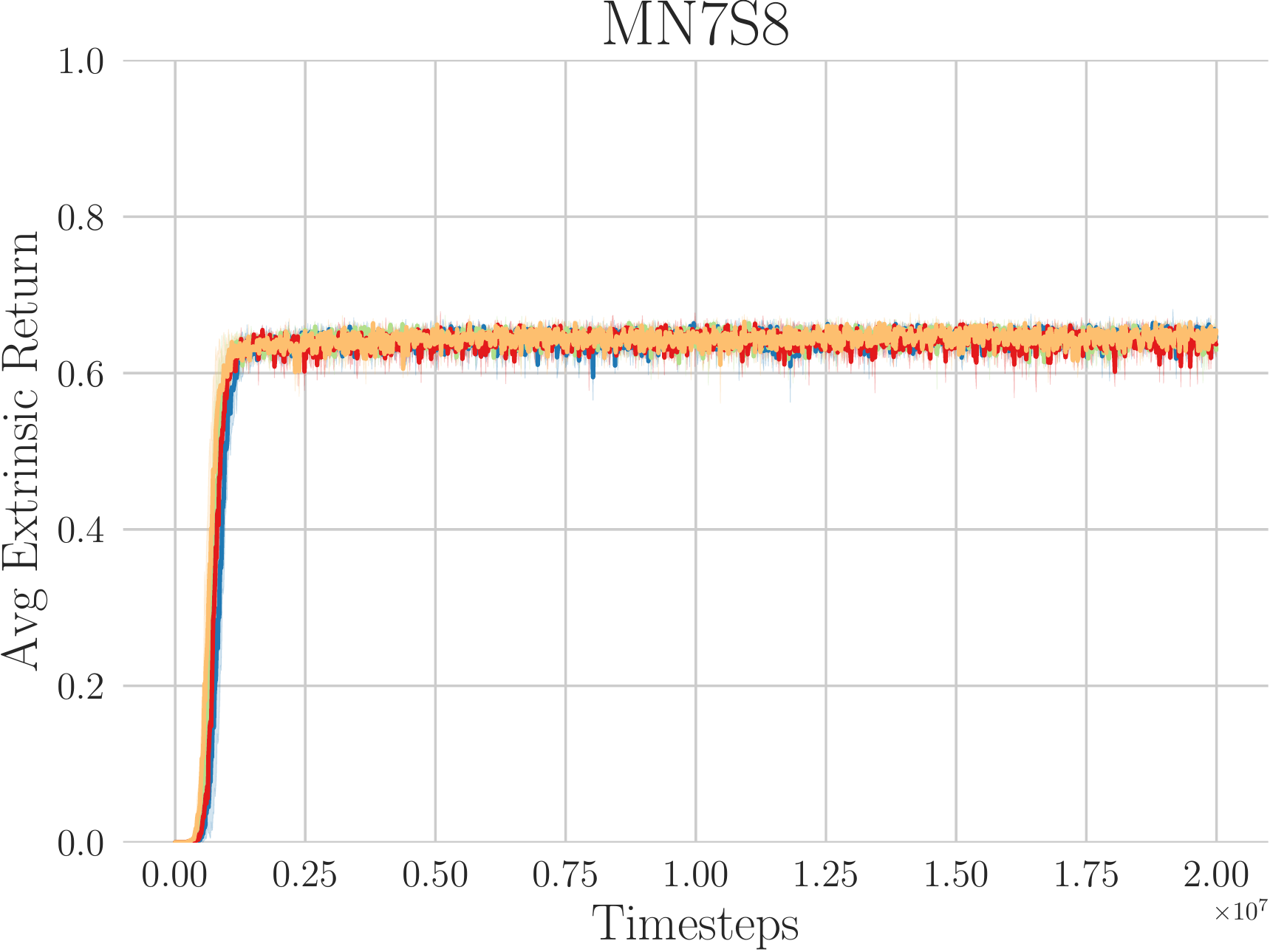}} 
    \subfloat{\includegraphics[width=0.5\columnwidth]{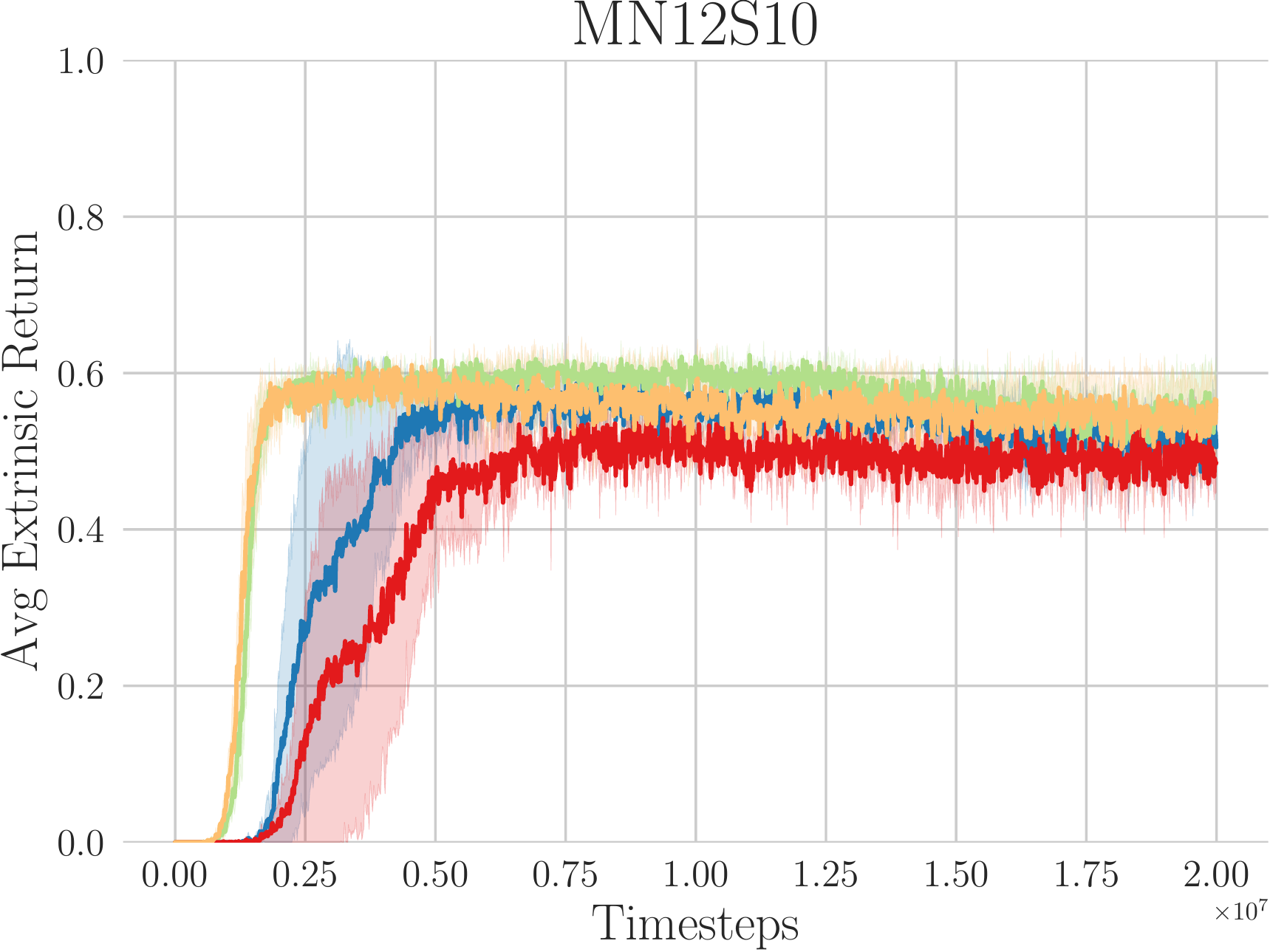}}
    \\
    \subfloat{\includegraphics[width=0.5\columnwidth]{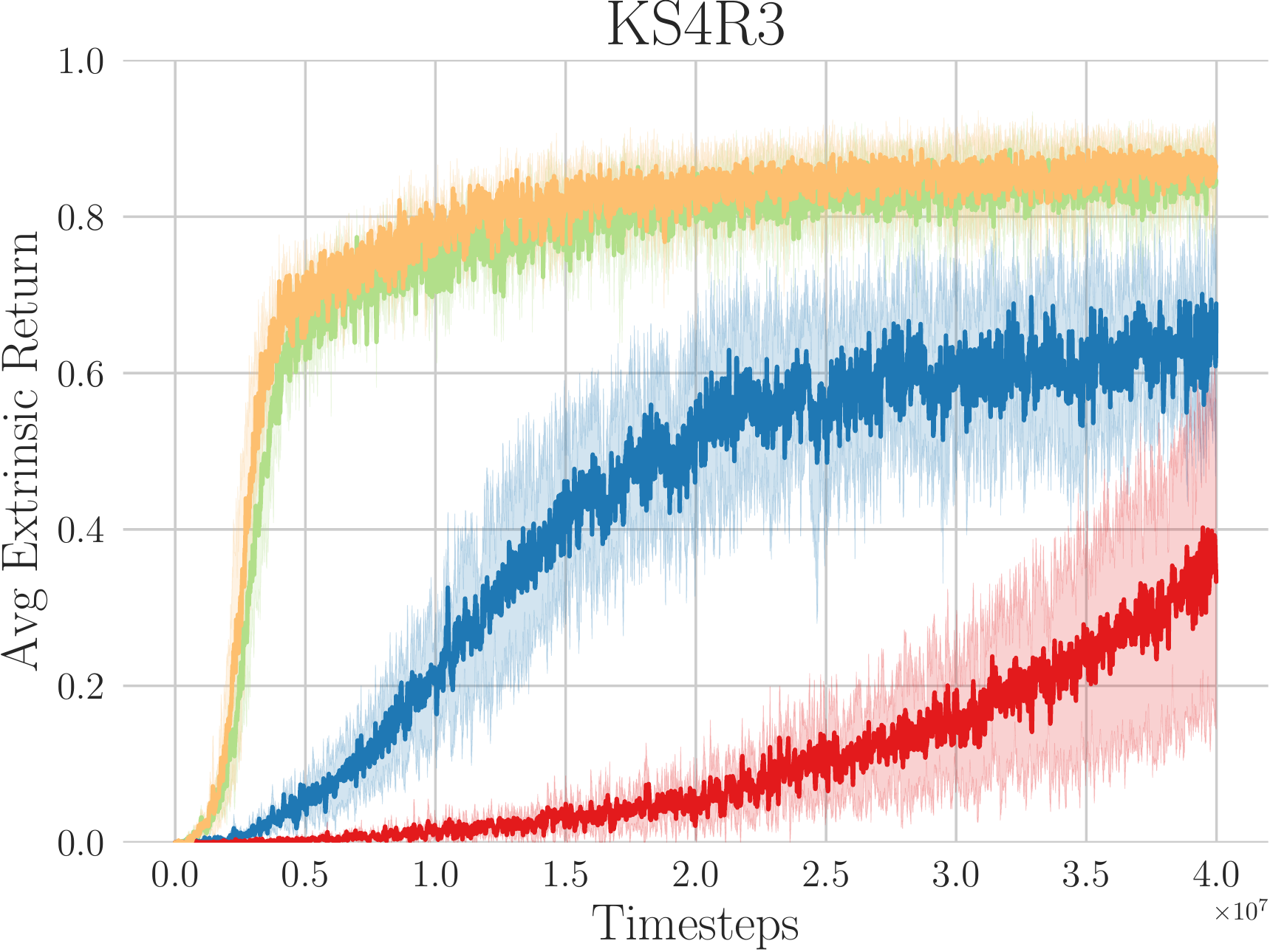}} 
    \subfloat{\includegraphics[width=0.5\columnwidth]{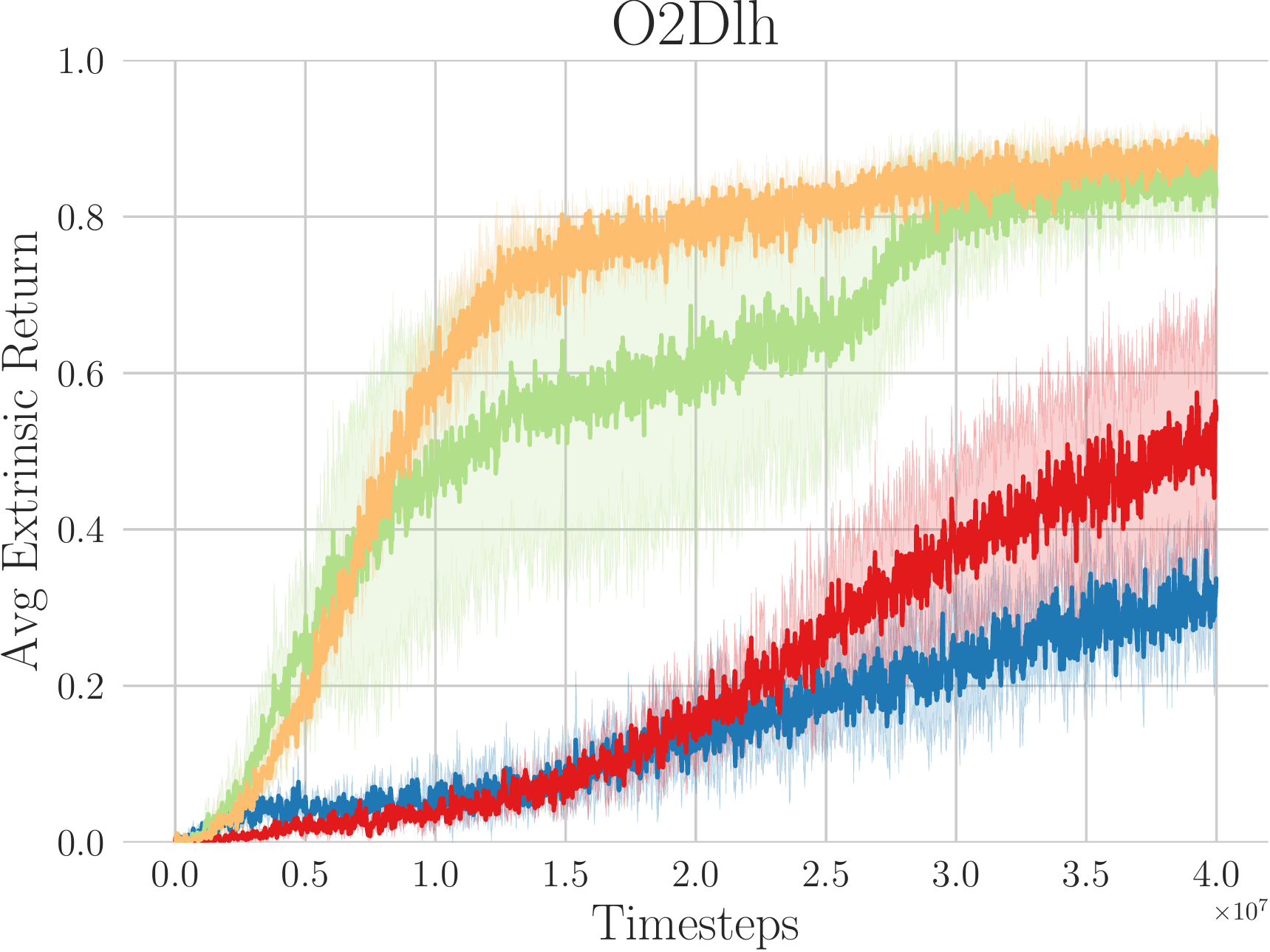}}
    \caption{Performance comparison of RAPID when combined with multiple IM methods (\emph{counts}, \emph{counts1st} and BeBold).}
    \label{fig:rapid+differentIMs}
\end{figure}

The results provided in Figure \ref{fig:rapid+differentIMs} suggest that there is a high relationship between what the agent can learn with IM (without self-IL) and what it actually does by combining them altogether. This can be regarded as a measure of the effectiveness of IM methods when implemented in isolation, where their base functionality of exploring is not wide-spread with the self-IL counterpart. At this point, by just inspecting the results reported in \cite{aa2022cdmake,zhang2021noveld}, it is clear that \emph{counts} is the worst method, followed by \emph{counts1st} and BeBold. Differences between \emph{counts1st} and BeBold are unclear: most of the contribution seems to be related to the episodic restriction part. However, going beyond the boundaries of already explored regions seems to be promising as well, as it yields better results when compared to RND with episodic restriction \cite{zhang2021noveld}. The same comparative performance between IM methods holds when combining them with the ranking replay strategy, where RAPID+\emph{counts} performs slightly better or equal to RAPID in isolation yet being the worst out of the 3 IM options. What is more, the choice of one IM strategy over another can actually deteriorate the performance of the agent, as observed in \texttt{KS4R3}. Nevertheless, when selecting  demonstrably good IM strategies, the agent combining self-IL+IM improves its performance even when it was not able to do it just with the IM strategy in isolation.

% 3. Introduce image 3 and variety of data shown.
\begin{figure*}[!b]
    \centering
    \subfloat{\includegraphics[width=1.0\textwidth]{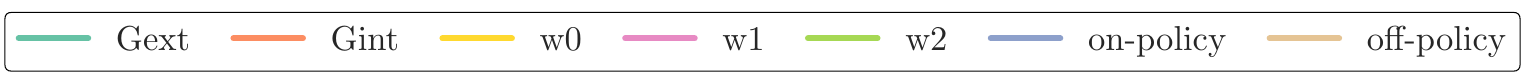}}
    \\
    \subfloat{\includegraphics[width=0.5\textwidth]{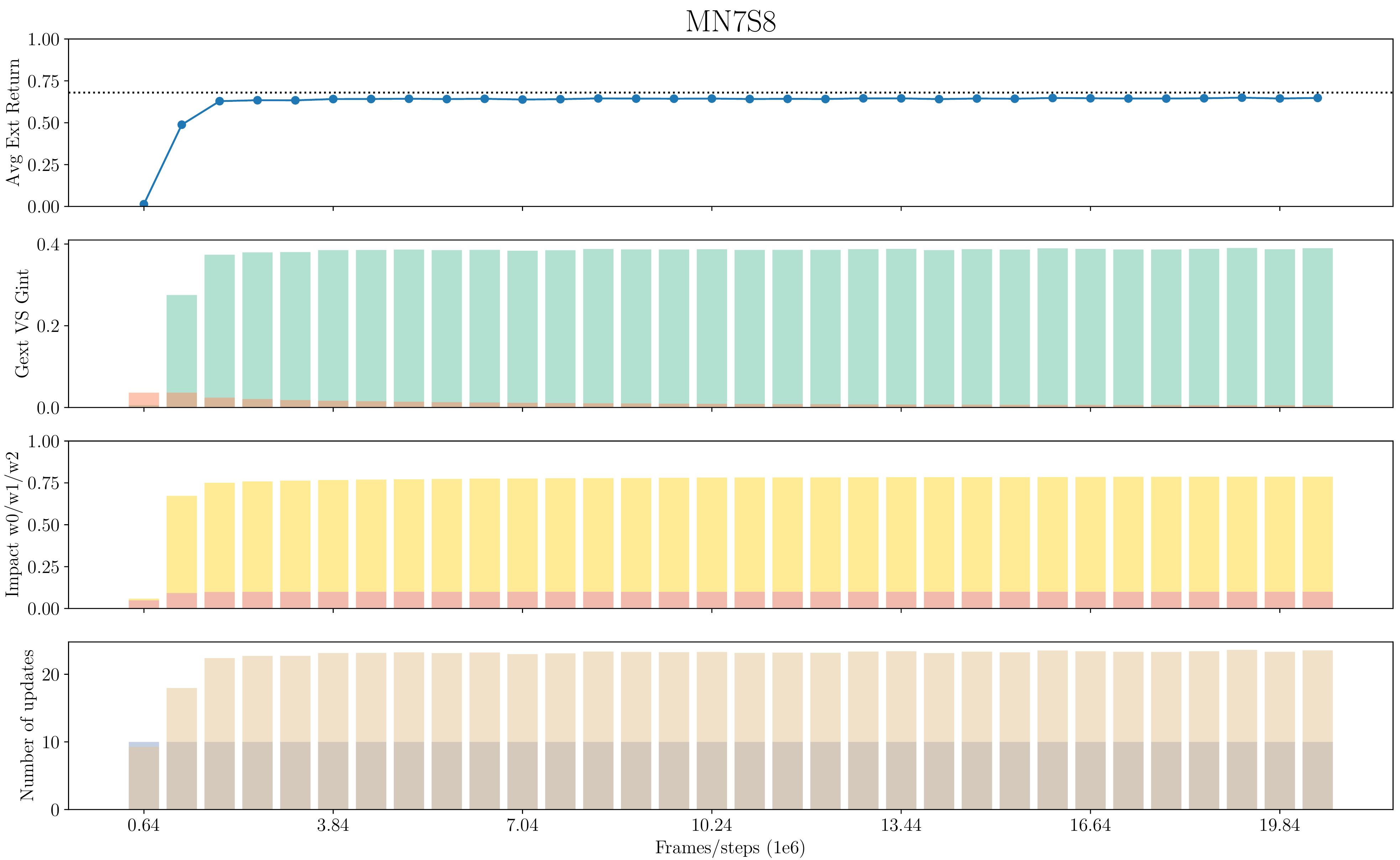}} 
    \subfloat{\includegraphics[width=0.5\textwidth]{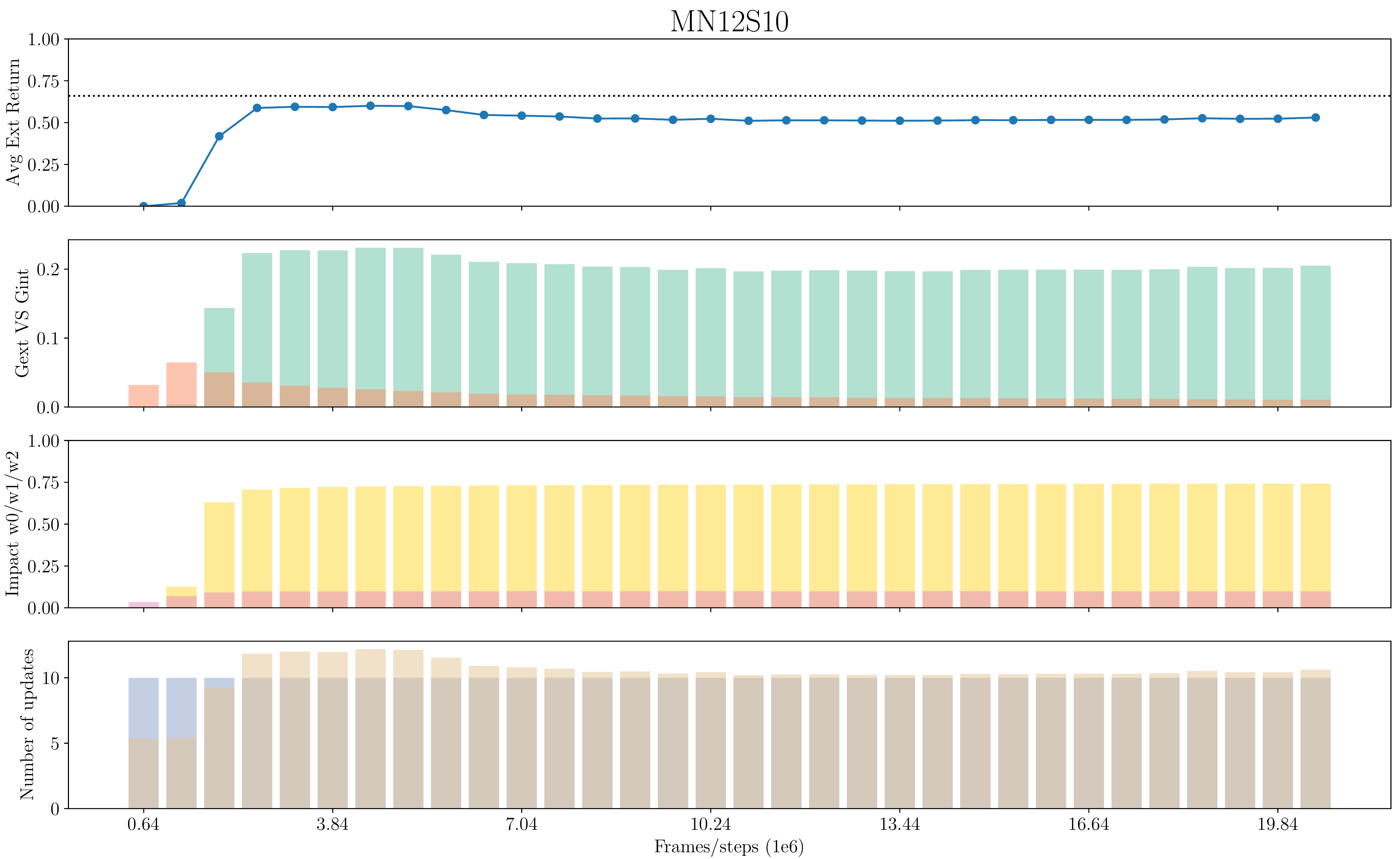}}
    \\
    \subfloat{\includegraphics[width=0.5\textwidth]{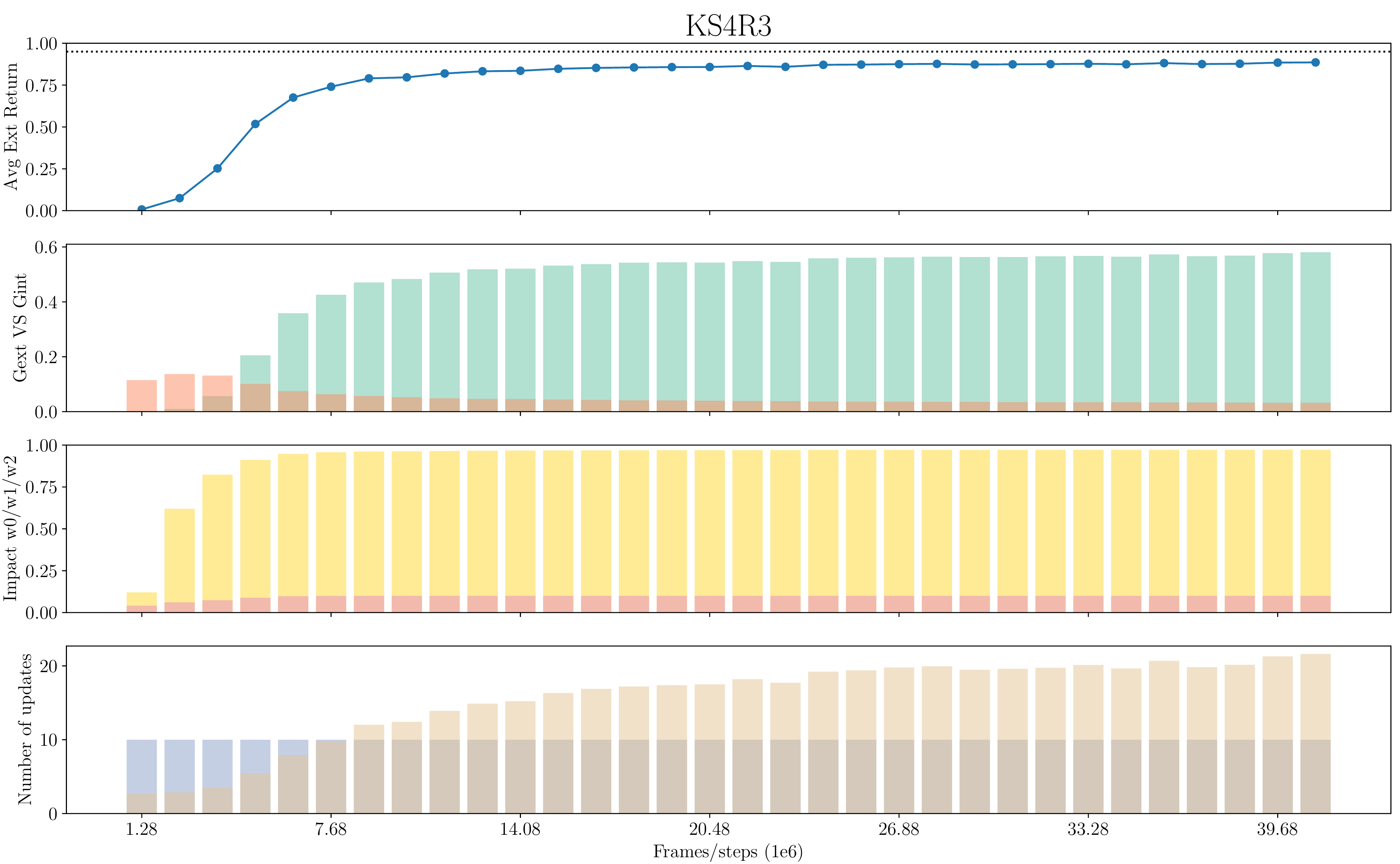}}
    \subfloat{\includegraphics[width=0.5\textwidth]{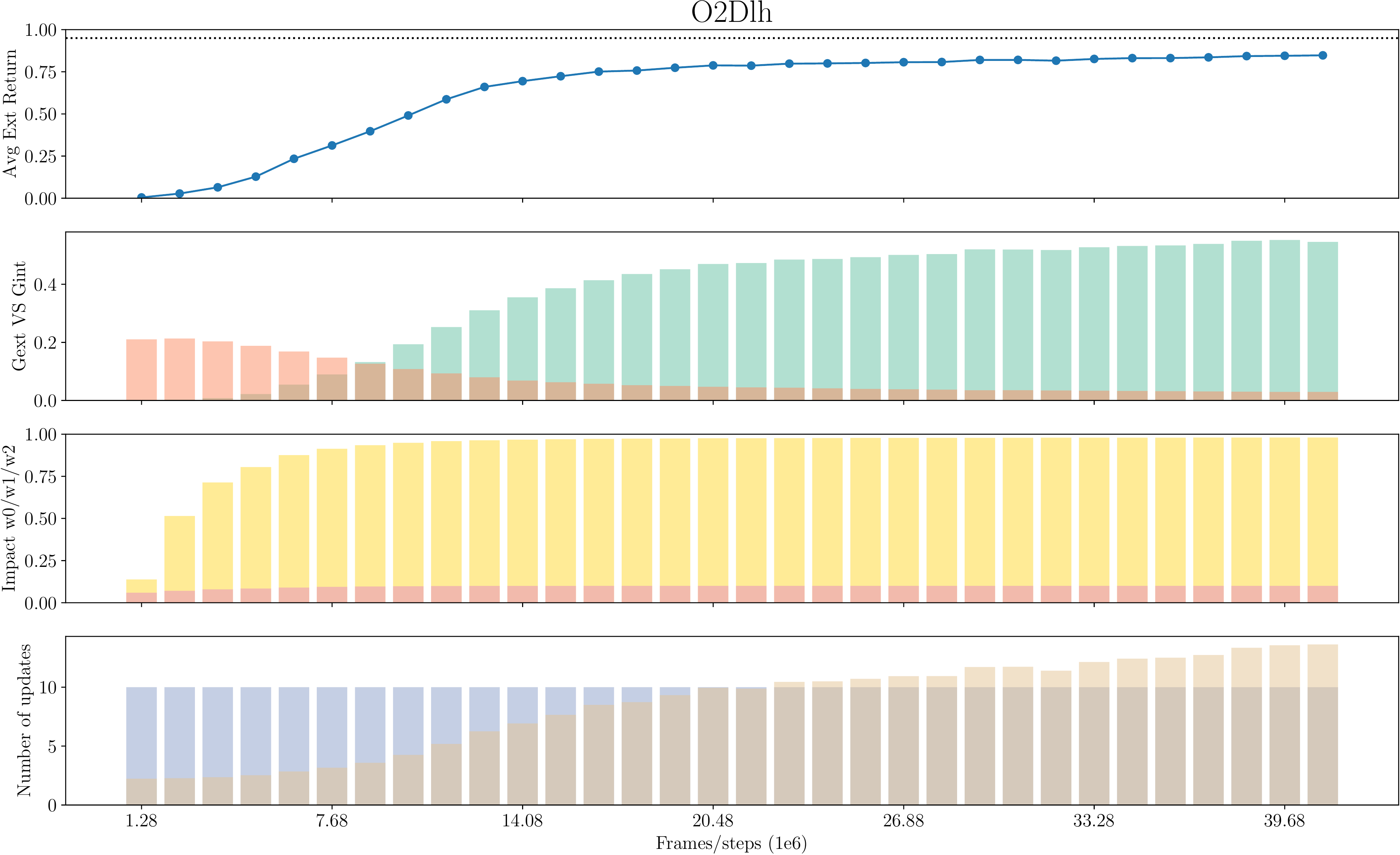}}
    \caption{Summary of the evolution of different critical values that impact the learning for a given seed in all the scenarios, using RAPID+BeBold. Plots in the first row denote the average extrinsic reward. Plots in the second row depict the difference between the discounted extrinsic ($Gext$) and intrinsic ($Gint$) returns used in the on-policy update (RL-loss). The third row of figures shows the influence of each component/score of the ranking buffer ($w0,w1,w2$) when sampling from its collected experiences. Finally, plots in the last row indicate the average number of off-policy updates per 10 on-policy updates (ratio of updates, $\xi$). All depicted data correspond to the average value in the given time slots.}
    \label{fig:scenarios_deep_analysis}
\end{figure*}
\textbf{Exploration-exploitation parameters in self-IL+IM:} by introducing IM into the on-policy loss, the agent has to deal with multiple objectives (exploration-exploitation) in various stages: 1) on-policy by balancing the extrinsic and intrinsic rewards; and 2) off-policy by keeping in the buffer the most promising experiences parameterized by the extrinsic, local and global scores. In this regard, Figure \ref{fig:scenarios_deep_analysis} depicts the evolution of some representative values concerning how the exploration is carried out during a training run. Initially $G_i(s)>G_e(s)$\footnote{Refers to the episodic discounted intrinsic and extrinsic returns calculated as described in Section \ref{sec:proposed}.} shows how the agent explores in the absence of extrinsic signals from the environment, yet eventually this gains more importance for the agent's ability to complete the task. Similarly, the impact of the $w_0$ score in Expression \eqref{eq:rapid} -- that promotes the exploitation of highly extrinsic rewarded episodes -- quickly increases. However, the selection criterion is also subject to the $w_1$ local score, which aims to maximize the diversity of observations inside the episode. In a much lower scale, the global scale $w_2$ also plays its role in the selection criterion, which can be helpful during the initial learning stages, when there are no success episodes to complete the task, and also to untie when two episodes require the same amount of steps for the task completion.

% 3. HABLAR SOBRE LA IMAGEN3, GRAFICA RETURNS, SCORES Y RATIOS
%Con el fin de poder entender mas en detalle como influyen los diferentes parametros de exploracion entre si, mostramos diferentes parametros de interes a lo largo del entrenamiento en cada escenario para un run concreto en la Fig \ref{fig:scenarios_deep_analysis}. Como era de esperar, al inicio el $G_i(s)$ es mayor que $G_e(s)$ lo que asegura cierta exploracion en la loss on-policy, pero eventualmente esto termina invirtiendose para promover sobre todo la parte extrinseca. Del mismo modo, el impacto de los diferentes scores que priorizan unos episodios sobre otros sigue un patron similar con w0, que es el relativo a priorizar la parte extrinseca. Curiosamente, debido a la parametrizacion de dichos scores, el componente explorativo global w2 apenas es relevante a excepcion en las etapas iniciales cuando apenas se han recodigo muestras con algun tipo de trayectoria con $G_e(s)>0$. Sin embargo, en el caso de tener dos niveles con el mismo score extrinseco y bonus local, este se utilizaria como ultimo caso. Asimismo, el bonus local que analiza la diversidad en las muestras recogidas se maximiza tan pronto como el agente es capaz de recolectar muestras que son optimas, donde la repeticion de observaciones es minimizada.

We proceed by evincing how the on-policy:off-policy ratio $\xi$ between the number of on-policy and off-policy updates change over the curse of training. On-policy optimization steps are executed once a trajectory has been finished, and it remains equal during the whole training.  Off-policy updates are instead applied once an episode finishes, which varies depending on the maximum steps per episode configured for each environment, and also on the optimality of the agent's policy at that moment. The decision to execute off-policy updates at the end of the episode was retrieved from the original paper where RAPID was proposed \cite{zha2021rank}. Such ratio $\xi$ can change from 1:1 to 1:3 in \texttt{MultiRoom} environments, and more dramatically in other scenarios like \texttt{KS4R3}, which initially implies a ratio of 4:1 and can evolve up to a 4:13 relation. In words, the off-policy loss can undergo a modification in its schedule that makes it update more than $10\times$ at its initial frequency (Table \ref{tab:table_ratios}). Such a balance has a critical importance in the agent's learning process, as it would turn to optimize what is stored in the buffer rather than what is actually experiencing (or vice versa). This generates in turn a big shift between both methods. In fact, in IL this ratio is usually balanced by either using a weight when combining both losses or by selecting the frequency update \cite{hester2018deep,sovrano2019combining}.

%Last but not least, mostramos el ratio o diferencia en el numero de updates on-policy vs off-policy. En el caso del primero, su ejecucion esta sujeta a terminar la coleccion de un rollout/trayectoria y es estatico durante todo el entrenamiento, mientras que el segundo es variable y depende de cuando se termina un episodio, lo cual depende de cada entorno en funcion del numero maximo de steps por episodio (cuando no es capaz de resolver) o el numero de pasos que le tome resolver el entorno (cuando lo resuelve)\footnote{El criterio de seleccion a fin de episodio esta basado en el criterio empleado por los autores de RAPID.}. Este ratio varia puede variar desde un 1:1 (on-policy vs off-policy) a 1:2 en entornos MiniGrid y puede llegar a cambiar varios ordenes de magnitud en los otros entornos, llendo de 10:2 a 10:20, lo que supone practicamente una diferencia de 10x en la frecuencia en la que se puede llegar a utilizar la loss off-policy! La determinacion del valor del ratio es algo llamativo y que creemos que influye en el desempeño de los resultados, ya que hara que la política se sesgue/haga caso más a una loss frente a otra.

\begin{table}[h]
    \begin{tabular}{cccccc}
    \toprule
    & & \texttt{MN7S8} & \texttt{MN12S10} & \texttt{KS4R3} & \texttt{O2Dlh} \\ \midrule
    \multicolumn{2}{c}{Max steps per episode}   & 140                        & 240                          & 480                        & 576                        \\ \midrule
    \multicolumn{2}{c}{Expected optimum steps}  & 50                         & 105                          & 37                         & 32                         \\ \midrule
    \multirow{2}{*}{$T=128$}  & Initial & 1:1                        & 2:1                          & 4:1                        & 5:1                        \\  
                        & Final & 1:3                        & 2:2                          & 4:13                       & 5:18                       \\ \midrule
    \multirow{2}{*}{$T=2048$} & Initial & 1:14                       & 2:17                         & 4:17                       & 5:18                       \\  
                            & Final & 1:40                       & 2:40                         & 4:216                      & 5:320                      \\ \bottomrule
    \end{tabular}
    \caption{On-policy vs off-policy ratios that can be achieved in each scenario when the supervised loss is subject to when the episode finishes. Each scenario has a different maximum number of steps (row 2) and also different expected number of optimal steps (row 3) (we include an estimation of the optimal steps as it differs from seed to seed). We show the expected initial ratios ($\xi$) when the agent cannot solve the task (rows 4 \& 6) and when it accomplishes the task via an estimated optimal policy (rows 5 \& 7). We also report those values when the rollout size is $T=128$ (rows 4-5) and $T=2048$ (rows 6-7).}
    \label{tab:table_ratios}
\end{table}

\begin{figure}[h]
    \centering
    \subfloat{\includegraphics[width=1.0\columnwidth]{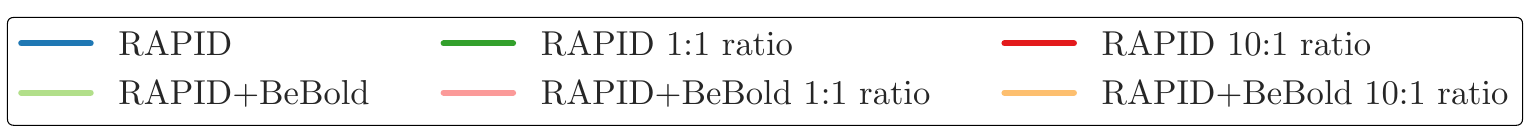}}
    \\
    \subfloat{\includegraphics[width=0.5\columnwidth]{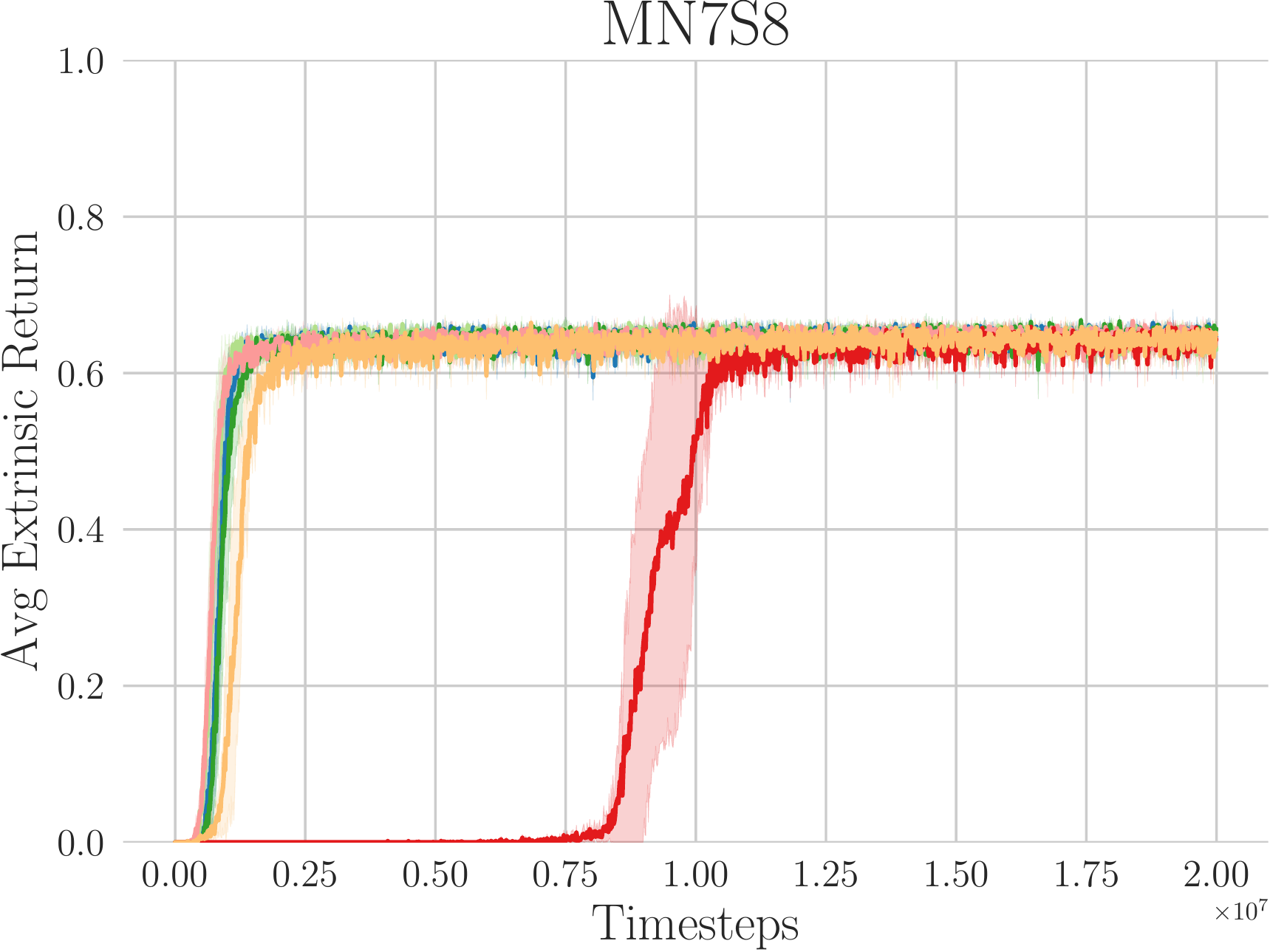}}
    \subfloat{\includegraphics[width=0.5\columnwidth]{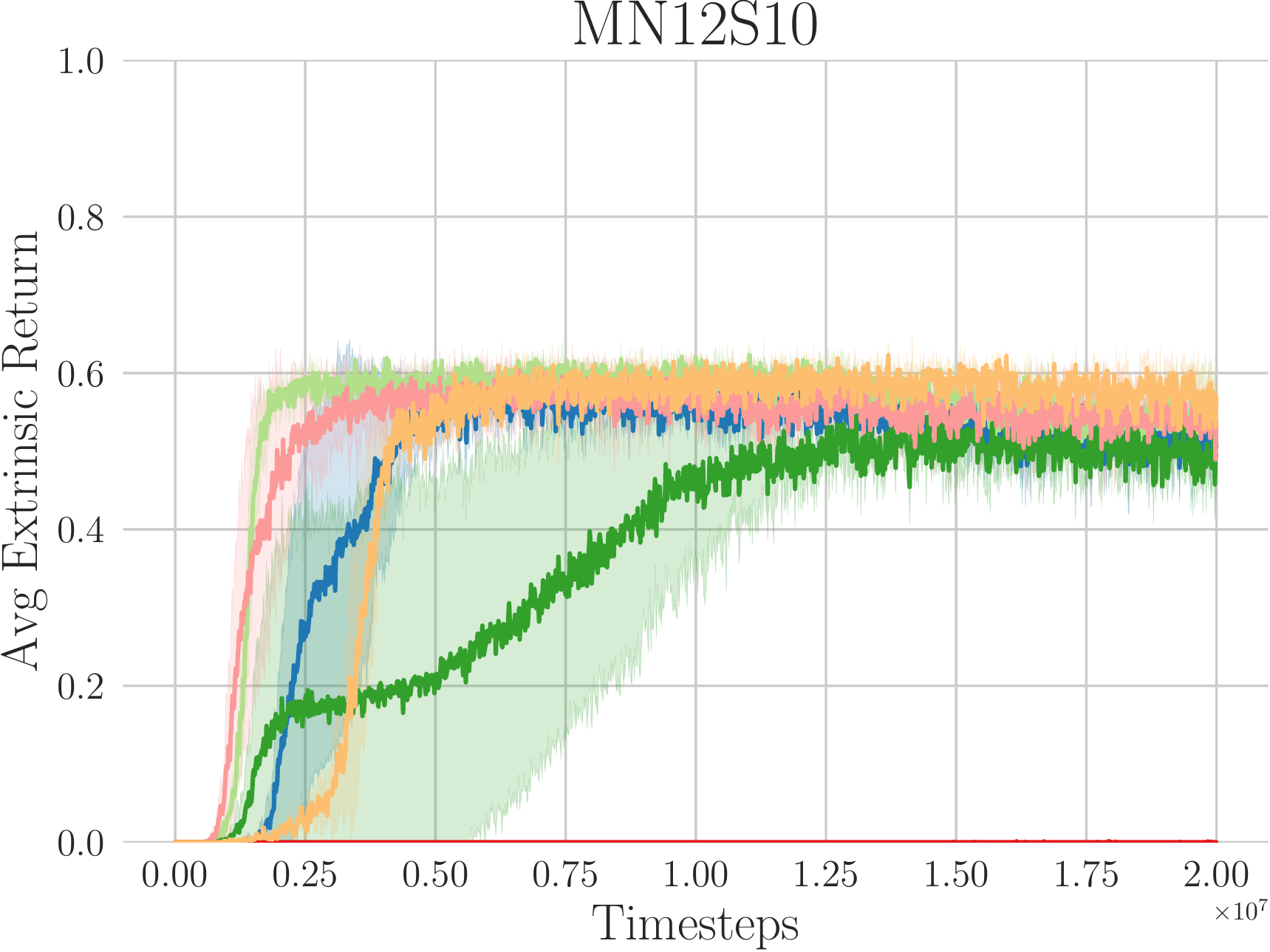}}
    \\
    \subfloat{\includegraphics[width=0.5\columnwidth]{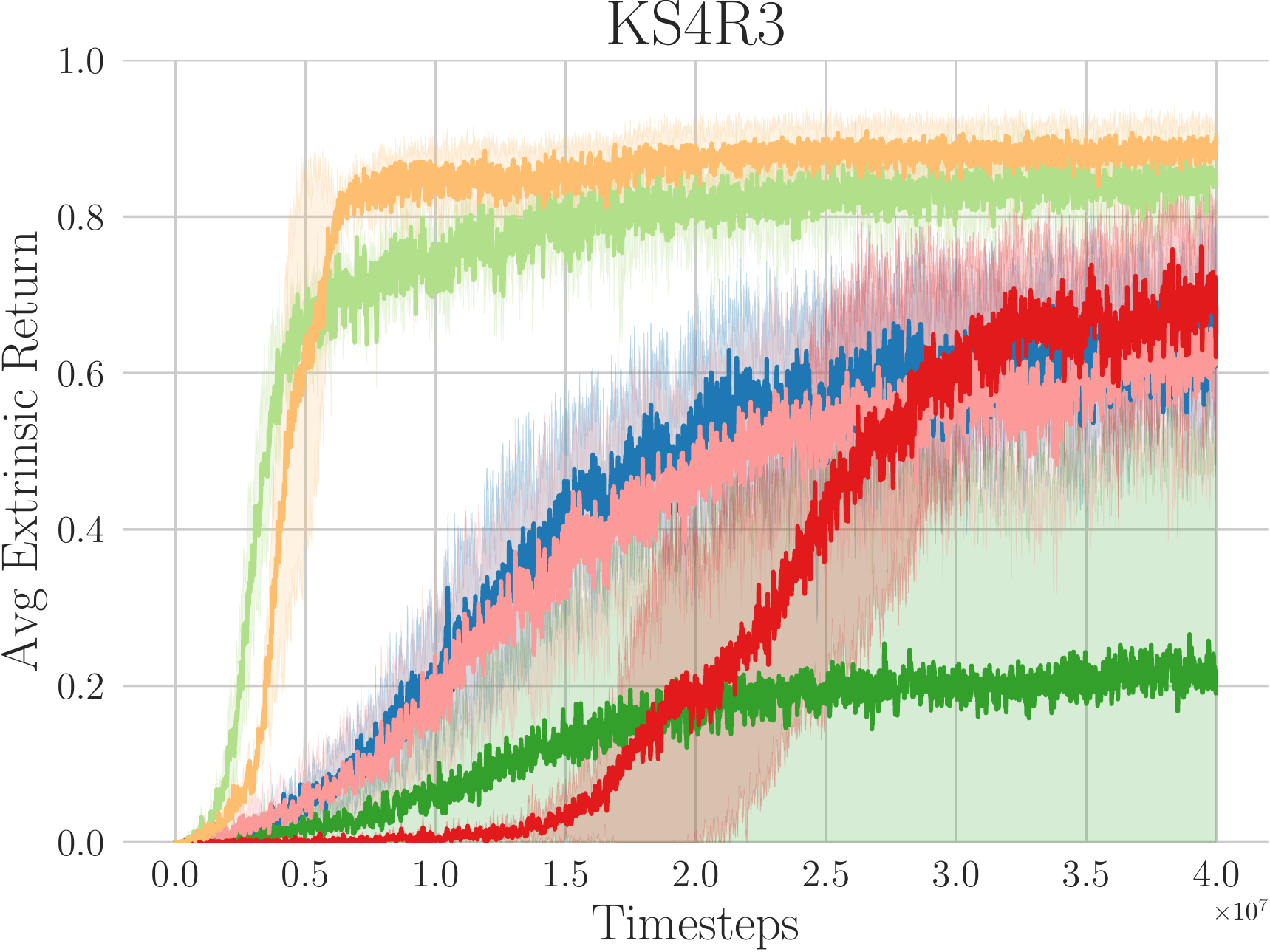}}
    \subfloat{\includegraphics[width=0.5\columnwidth]{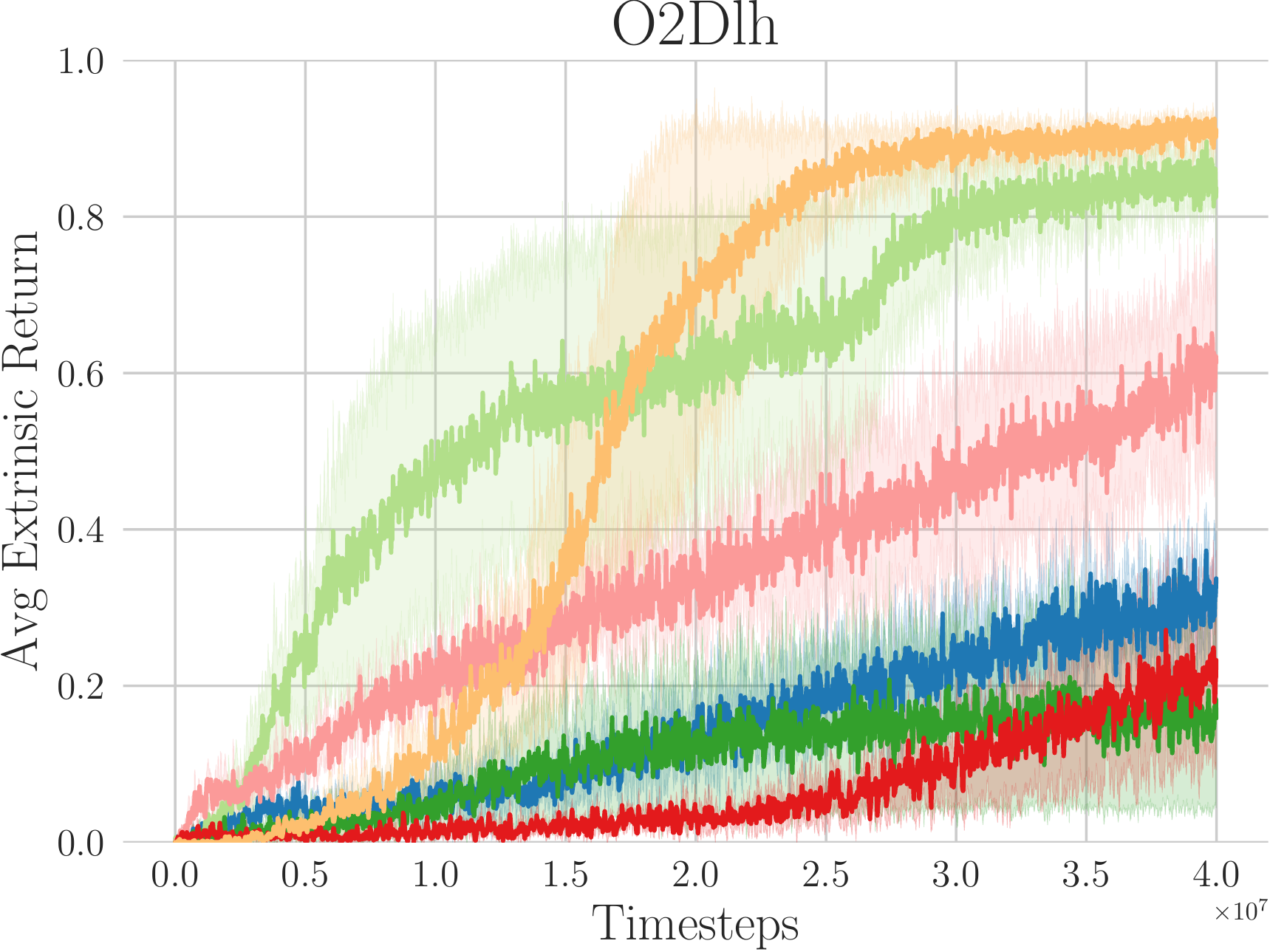}}
    \caption{Results over multiple procedurally-generated MiniGrid hard exploration environments using different ratios between on-policy (PPO) and off-policy (RAPID) updates. The default RAPID approach has a dynamic update ratio, by which it executes an optimization step every time an episode finishes (see Table \ref{tab:table_ratios}).}
    \label{fig:core+ratios}
\end{figure}

% 4. DIFERENCIA DE RATIOS AL USARSE BATCH 32
\vspace{2mm}
\textbf{Scheduling the self-IL updates}. To shed further light on the importance of the aforementioned ratio, we now fix the off-policy loss to be constant and subject directly to the on-policy updates. We then analyze how the performance varies under several values for this ratio. Figure \ref{fig:core+ratios} summarizes the results obtained for this study. In the family of \texttt{MultiRoom} scenarios, the agent is very sensitive to a reduction of the frequency of the off-policy updates, which can eventually make the agent fail when increasing their complexity (10:1 in \texttt{MN12S10}). Contrarily, in \texttt{KS4R3} the original adopted schema with a ratio of 4:1 performs much better than a more frequent update of the off-policy part (1:1). This fact is also observed when using a more conservative ratio of 10:1, suggesting that, although a higher off-policy update frequency can be beneficial at initial stages to bootstrap the learning process in hard exploration tasks, it can eventually degrade the learned knowledge in the long term. These conclusions can also be inferred when using BeBold, but with a better sample-efficiency and optimal solutions. Similar conclusions hold when analyzing \texttt{O2Dlh}.

%Partiendo de la importancia que puede llegar a tener el ratio entre updates, hemos decidido establecer un ratio fijo entre ambas losses haciendo que la off-policy dependa directamente sobre su homolaga on-policy. Para analizar el impacto de diferentes ratios, hemos seleccionado diferentes valores tal y como se muestra en la Fig \ref{fig:core+ratios}. En los escenarios MultiRoom el agente es muy sensible a decrementar el numero de veces con las que se utiliza la supervised loss, empeorando varias ordenes de magnitud su velocidad de convergencia e incluso fallando en el caso de 10:1. Por otro lado, en KS4R3 encontramos un comportamiento completamente opuesto, donde el valor por defecto (inicialmente 4:1) funciona mejor que utilizando un ratio donde se fomenta más el uso de la supervised loss (1:1). Esto se alinea tambien cuando se emplea el ratio 10:1, lo que sugiere que utilizar un ratio igual en ambos casos puede llegar a ser beneficioso inicialmente pero que después es mejor dejar que prevalezca lo que sugiere directamente la parte on-policy. Estas mismas conclusiones son extrapolables a cuando se aplica BeBold, pero con mejora en sample efficiency y optimality. Asimismo, en O2Dlh las conclusiones son del mismo estilo a las explicadas anteriormente en KS4R3.

%JAVI 12/07/2022
% 5. AUMENTAMOS TAMAÑO DE BATCH A 2048
\vspace{2mm}
\textbf{Addressing inter-episode variance}. So far, the selected ratio seems to be decisive for the success and sample-efficiency of the training process. However, the obtained outcomes are very noisy and barely close to optimal results, which can be due to one of the two losses being unstable. While the seminal work presenting RAPID used PPO with a rollout size\footnote{The rollout size is directly related with the number and minibatch size. The increment of the first implies that the minibatch size is also incremented (for the same number of minibatches).} of $nstep=T=128$, other similar works considering the same environment used a larger time horizon $nstep=T=2048$, with better and more stable results \cite{aa2022cdmake,flet2021adversarially}. As mentioned in Section \ref{subsec:exp_env}, each episode is configured differently depending on the selected seed. Consequently, by training the agent with less episodes in a single update, it might get biased to learn specific features present in that subset of episodes, rather than getting the required high level skills to solve the desired task in the whole possible episode distribution. Hence, the increment of the rollout size implies that the agent will be trained (in the on-policy update) with a larger set of episodes (see Table \ref{tab:table_ratios} to check for episode lengths) which forces the algorithm to extract generalizable knowledge in this wider set of slightly different environments, preventing from a by-heart learning.
Furthermore, this also reduces the variance of the on-policy updates through the neural network, as the minibatch size itself will be larger. However, the agent will perform less optimization steps during the training process for the same amount of steps/frames. 

\begin{figure}[h]
    \centering
    \subfloat{\includegraphics[width=1.0\columnwidth]{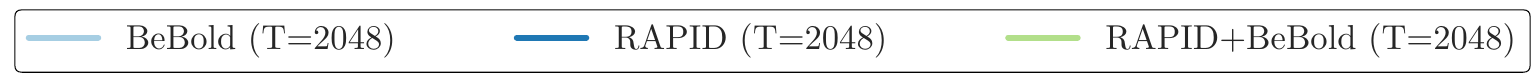}}
    \\
    \subfloat{\includegraphics[width=0.5\columnwidth]{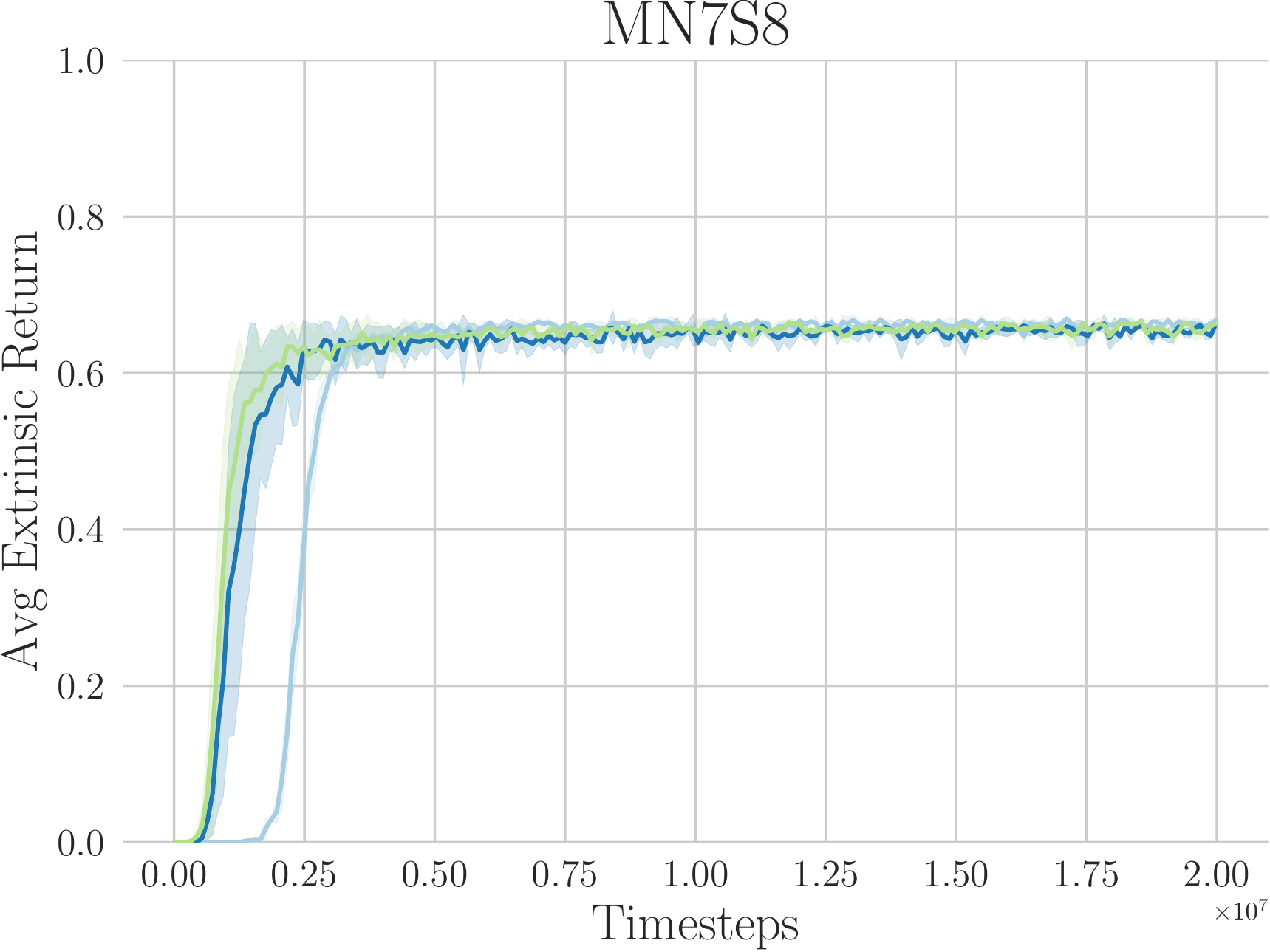}} 
    \subfloat{\includegraphics[width=0.5\columnwidth]{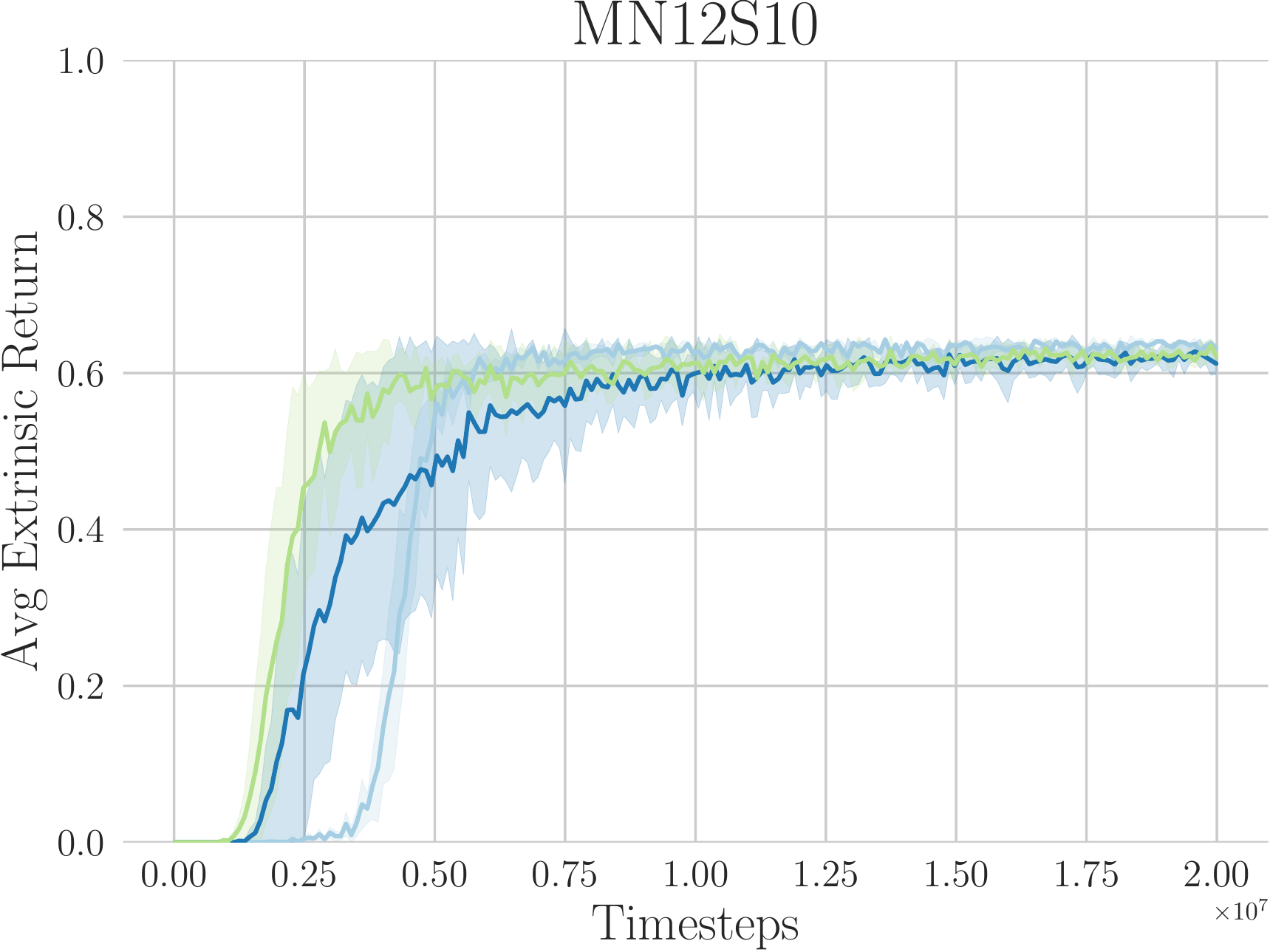}}
    \\
    \subfloat{\includegraphics[width=0.5\columnwidth]{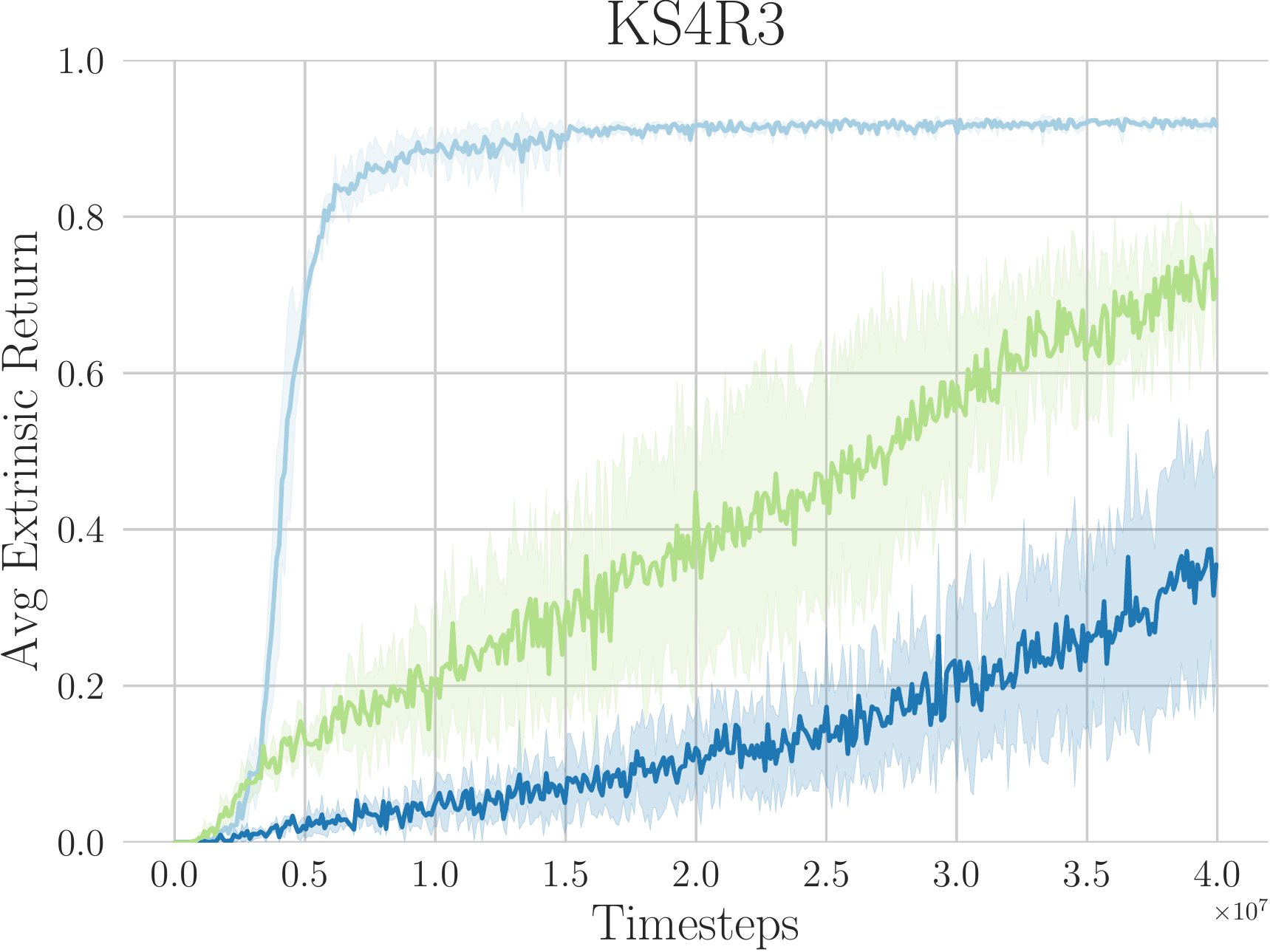}} 
    \subfloat{\includegraphics[width=0.5\columnwidth]{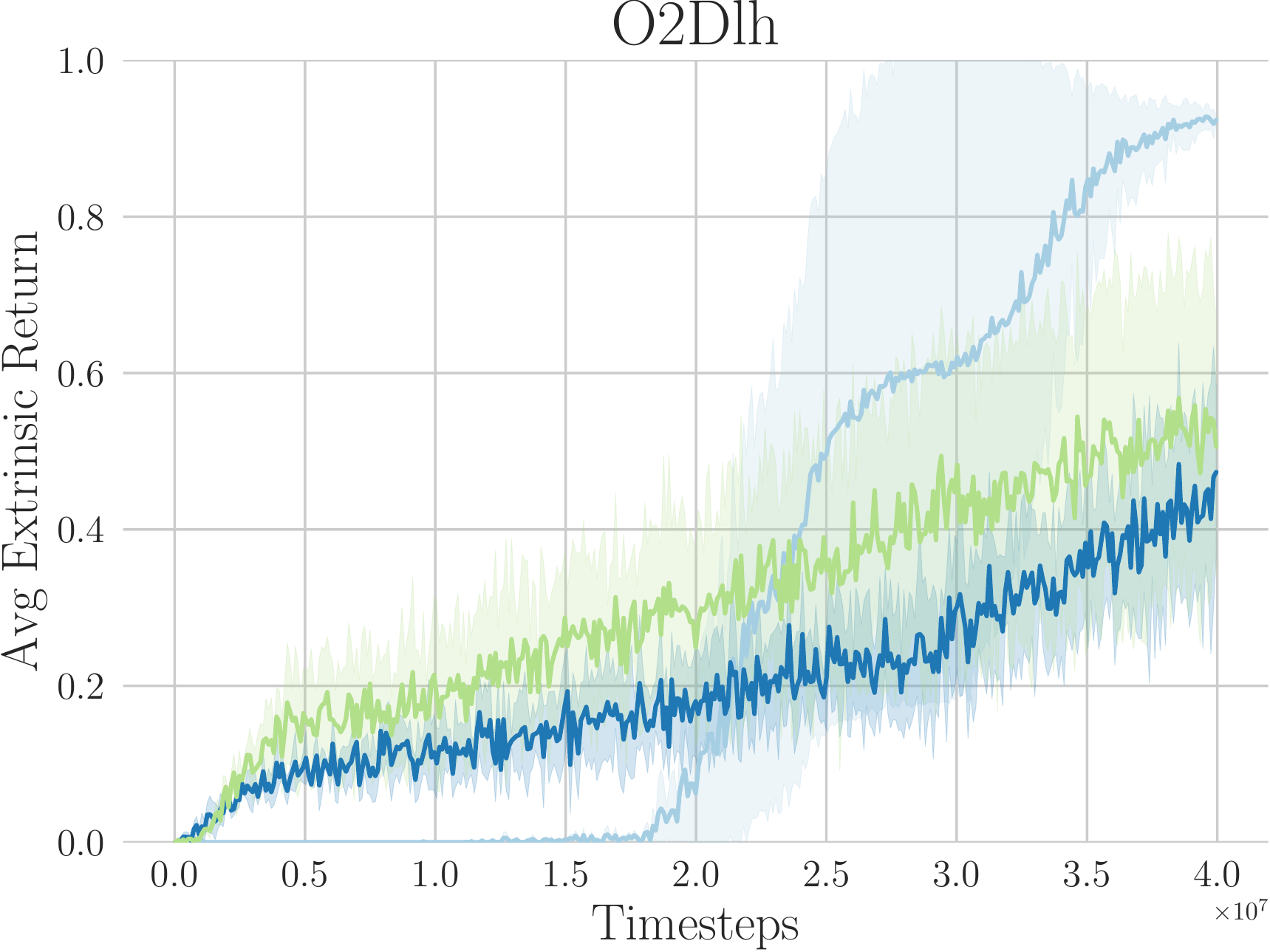}}
    \caption{Results on multiple hard exploration procedurally-generated environments in MiniGrid when increasing the time horizon up to 2048 in on-policy (RL-loss) updates. Off-policy (supervised/imitation) updates remain with fixed batch size of 256.}
    \label{fig:core_results_t2048}
\end{figure}

In this context, \emph{how does the use of larger rollout size impact on the on-policy update regarding the performance and the stabilization of the learned knowledge?} The answer to this question can be found by analyzing Figure \ref{fig:core_results_t2048}. The on-policy update is substantially improved, which can be seen in how BeBold performs without being corrupted by off-policy updates, being such IM solution able to solve all the environments with the expected optimal steps and obtaining the best result in both \texttt{KS4R3} and \texttt{O2Dlh}. Contrarily, RAPID performs worse, and its contribution when combined with BeBold is also not as good as it has been observed in the previous analysis. 

\begin{figure}[h]
    \centering
    \subfloat{\includegraphics[width=1.0\columnwidth]{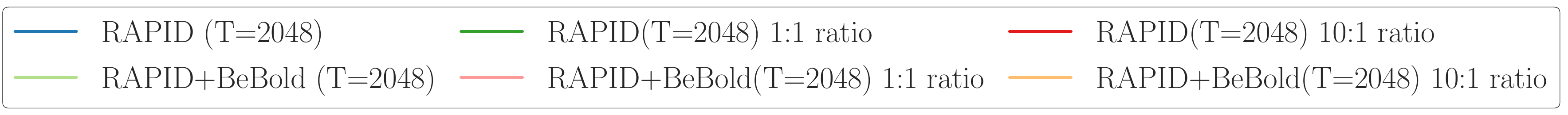}}
    \\
    \subfloat{\includegraphics[width=0.5\columnwidth]{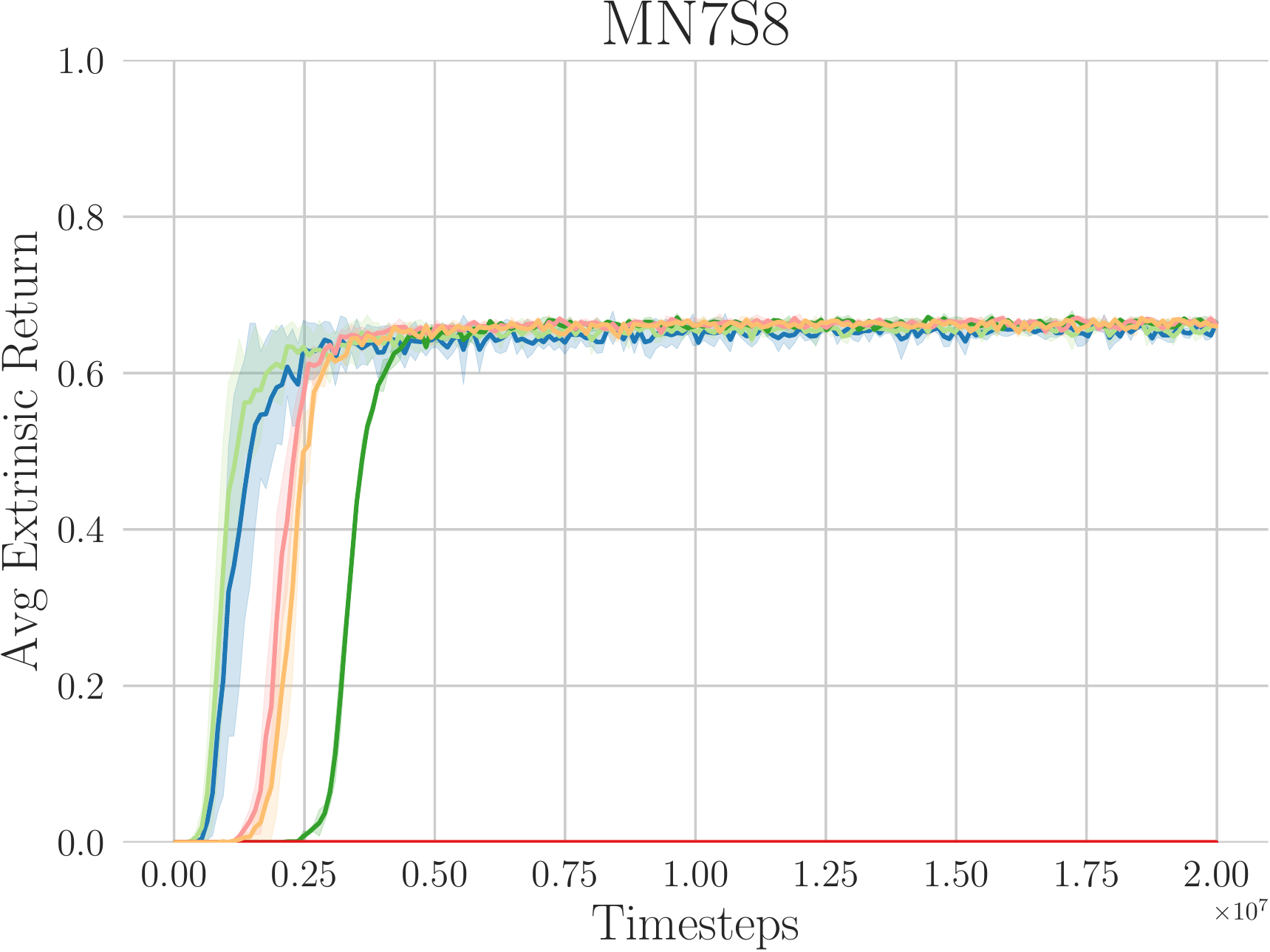}} 
    \subfloat{\includegraphics[width=0.5\columnwidth]{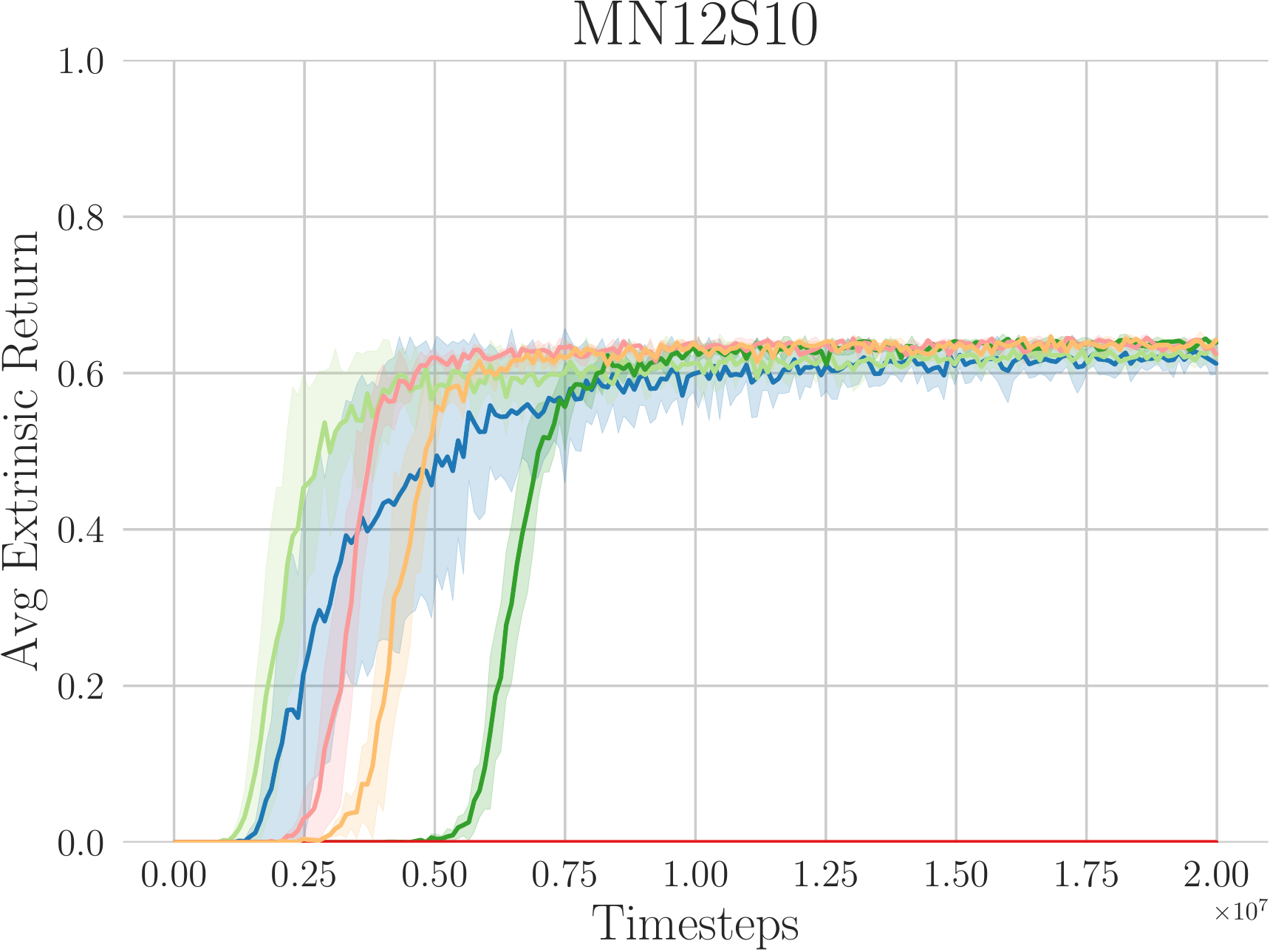}}
    \\
    \subfloat{\includegraphics[width=0.5\columnwidth]{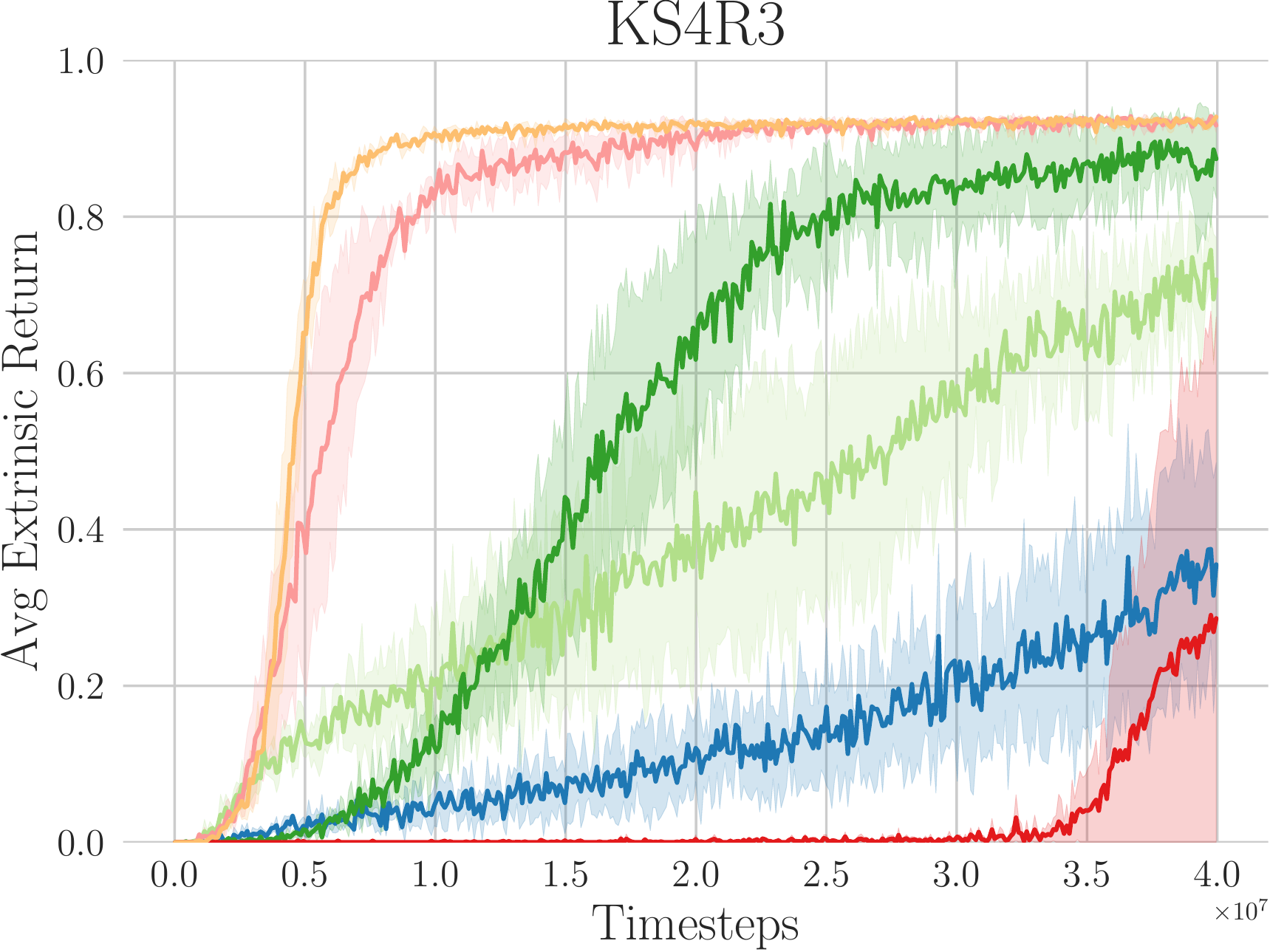}}
    \subfloat{\includegraphics[width=0.5\columnwidth]{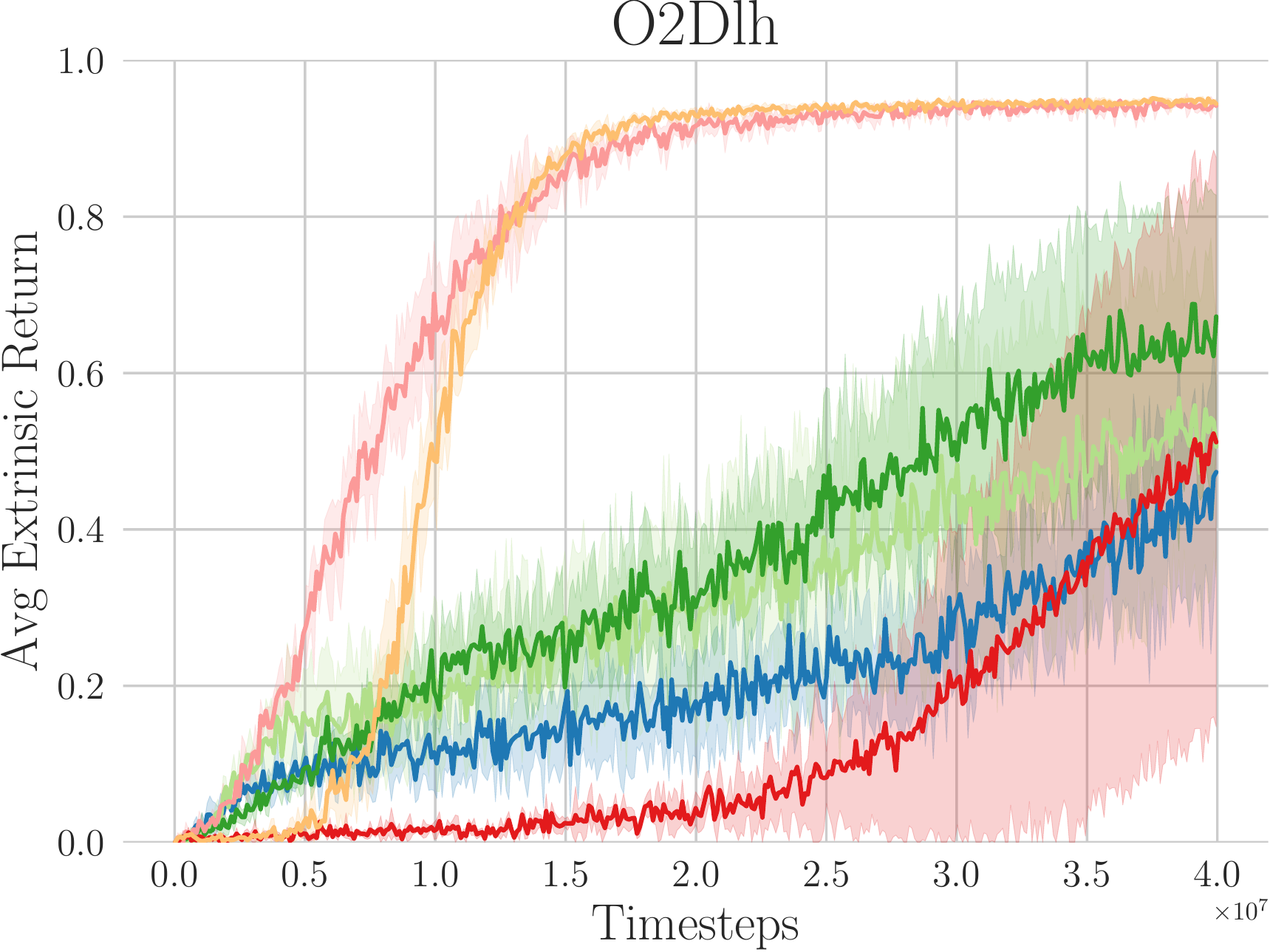}}
    \\
    \caption{Same interpretation of ratios as in Fig \ref{fig:core+ratios} but when increasing the rollout size up to 2048. Recall the default ratio is dynamically adjusted based on when the episode finishes (see Table \ref{tab:table_ratios}).}
    \label{fig:core+ratios+t2048}
\end{figure}

% 6. DIFERENCIA DE RATIOS AL USARSE BATCH 2048
The reason for such bad results is also connected to what we have previously highlighted: the ratio $\xi$. By increasing the rollout size and making the off-policy updates be subject to the episode completion, the off-policy loss grows up to be $14\times$, $8\times$, $4\times$ and $4\times$ more frequent than the on-policy counterpart in \texttt{MN7S8}, \texttt{MN12S10}, \texttt{KS4R3} and \texttt{O2Dlh}, respectively, just at the start of the training process (Table \ref{tab:table_ratios}). As we have already seen in Figure \ref{fig:core+ratios}, those kind of ratios does not necessarily mean to foster a better learning. Thus, when adjusting the ratio again with the new rollout size the performance of both RAPID and RAPID+BeBold drastically changes, as seen in Figure \ref{fig:core+ratios+t2048}. In first instance, it can be seen a better sample-efficiency when using a more conservative ratio (1:1) in both \texttt{KS4R3} and \texttt{O2Dlh}. This also happens when decreasing the off-policy updates up to a 10:1 ratio, where the convergence speed can be affected although it manages to achieve the optimal policy in less steps (the 1:1 ratio struggles more to finally achieve it). In contrast, when using the criteria of applying those updates at the end of the episode, which corresponds with approximately a 1:4 ratio initially in \texttt{KS4R3} and \texttt{O2Dlh} (Table \ref{tab:table_ratios}), the results get worse, just surpassed by the BeBold approach. These behaviors strengthen our claim: \emph{the off-policy loss can help to improve the learning process, although using it in excess can be counter-productive}. This is related to what is actually aiming to replay and if it is worth the value that update for the agent at that moment. Concerning \texttt{MultiRoom} environments, increasing the number of off-policy updates seems to be a good strategy, which is difficult to be beaten even by other state-of-the-art solutions. In fact, decreasing the frequency of the replayed experiences has a negative impact that can lead the agent to not learning in the absence of intrinsic rewards.

\subsection{Discussion} \label{subsec:limitations}
\subsubsection{Environment dependant requirements}
% 1.PORQUE FUNCIONA MEJOR MULTIROOM - REQUISITOS DE APRENDIZAJE DE CADA ENTORNO
So far we have seen outstanding results in either \cite{zha2021rank} and the current work for \texttt{MultiRoom}, although such benefits are not so clear in other scenarios. We hypothesize that this due to the understandings and latent knowledge that is needed to interact with the environment to accomplish the task; this is, \emph{what the agent actually has to learn in each environment to accomplish the task}. On the one hand, in \texttt{MultiRoom}'s environments the agent independently of the destination has always to move forward until finding a door (any color), open it and continue until reaching the goal. Hence, the agent has to discover that the way to solve the task is to find the next door as soon as possible and open it with the consequent action, which may well be inferred and faster learnt when exploiting past episodes as all that information does not indeed change from one level to other. On the other hand, in \texttt{KS4R3} and \texttt{O2Dlh} the agent has to learn how to interact with multiple objects (door can be closed or locked), the relation between objects and their utility (key is used to open lock doors) and the information given by the colors (the key of a given color only unlocks the door of that same color). Furthermore, the location of doors, keys and also the destination change from level to level so exploiting past good episodes do not necessarily mean to be the best strategy, because the stored episodes might be biased to certain patterns that can bias the agent's intuition (i.e. blue doors represent the best strategy to achieve the goal) and hinder what it has to actually learn.

Aside from the environment requirements, it is not straightforward to determine \emph{what episodes to replay and when to do it}, as what the agent learns changes over the curse of the training. By replaying just a certain portion of all the possible levels we highlight two main drawbacks:

\subsubsection{Intra/Inter Level Diversity}
% 2.QUE COÑO SE ESTA PRIORIZANDO... TIENE SENTIDO? ASEGURAR DIVERSIDAD
% diversity
The first is related to the \emph{diversity across episodes}. Diversity in the buffer is desired when the environment evolves during the training, thus maximizing the probability of having something useful in the buffer for the agent's learning at the moment. Rapid hyper-parameterization selection will be decisive for the buffer configuration\footnote{In this work, for the sake of comparison, parameters are set with the same values as in the original paper \cite{zha2021rank}.} prioritizing either the emulation of past levels with high intra-episodic diversity of states through local bonus or long-term exploration via global bonus. However, none of them was originally designed to control the diversity across different levels (inter-level). In addition, being these scored weighted statically and with no prior insights about what the agent actually manages to solve makes it impossible to assure that the buffer contents will be advantageous to learn convenient generalized knowledge. As a consequence, the agent would be forced to imitate levels that do not hold any guarantees to be useful when exploiting generalization.
%With RAPID we aim at maximizing the diversity of states/observations within the same episode/level (intra episode diversity), which encourages to prioritize the replay of those levels that counts on a large variety of states due to the local bonus\footnote{Independently of the designed scores, the results will also be subject to the selected weights given to each of them. In our case, we use equal to those used in RAPID \cite{zha2021rank} (see Section \ref{subsec:baselines_hyperparams}).}. In spite of using the global bonus to encourage long-term exploration during training, it does not directly address the diversity across different levels as the exploration patterns can change from one environment to other and do not necessarily hold any significant diversity difference\footnote{The concept of diversity can be subjective and it is difficult to be defined; when do we consider if a level is part of the whole distribution or if it is an outlier? For example, two levels can be considered very similar because having the objects positioned in the same positions yet having different colors; or vice versa, objects of the same color but differently positioned.}.As a consequence, the agent will be forced to imitate levels that do not hold any guarantees to be useful when exploiting generalization.

\subsubsection{Suboptimal demonstration replay}
\begin{figure}[h]
    \centering
    \subfloat{\includegraphics[width=1.0\columnwidth]{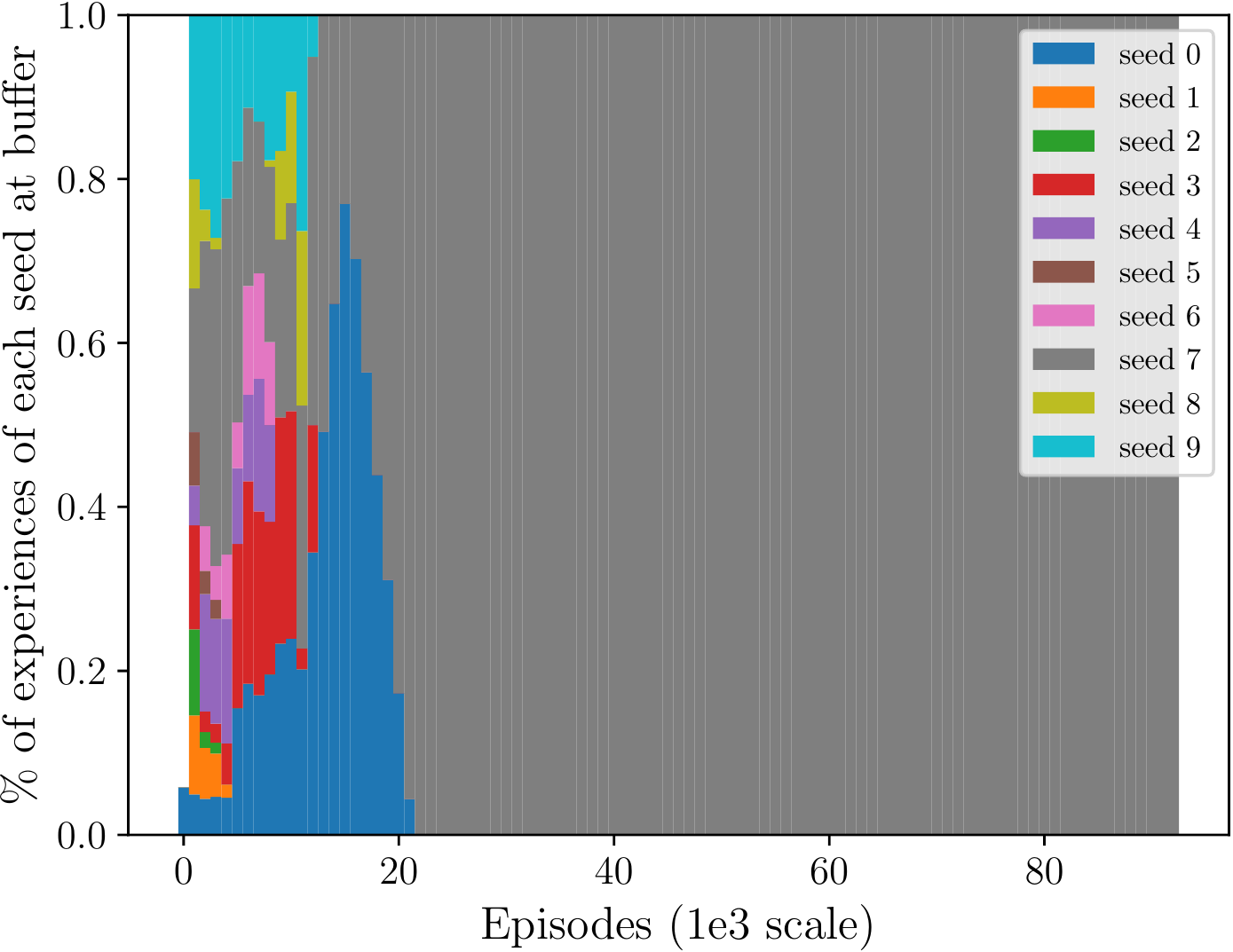}} 
    \caption{Experiences stored in the ranking buffer related to each level/seed during the training in \texttt{O2Dlh} when fixing the number of seeds to 10. Across those levels we have an average optimal number of steps of 18, being the level with seed 7 the one with a configuration that requires the minimum steps to solve the environment with just 13 steps. After 20,000 training episodes, the agent just stores and imitates greedily the level corresponding to seed 7.}
    \label{fig:o2dlh_seeds_fixed_buffer}
\end{figure}

% 3.PROBLEMA CON LAS SEEDS Y EL EXTRINSIC REWARD - NO EVALUA LA OPTIMALIDAD!
%seeds
The second is referred to \emph{how the sparse extrinsic reward is calculated in procedurally-generated environments}. Such function is commonly designed taking into account the number of steps until achieving the goal; particularly in MiniGrid it is calculated as $R(t) = 1 - 0.9\cdot t/t_{max}$, being $t$ the number of steps to solve the environment and $t_{max}$ the maximum number of steps that can be taken in an episode before being reset. Nevertheless, the levels have a minimum number of optimal steps based on the sorted configuration. Hence, the same reward at different levels would not necessarily have to represent the same optimality. Consequently, \emph{the agent will be more prone to learn from suboptimal demonstrations of those "easy" configured levels} just because solving them suboptimally takes less steps than doing it optimally in other levels. Furthermore, \emph{the agent can become greedy and bias its learning to those "easy" levels that do not necessarily represent the whole level distribution of the given task and fail in generalization}. What is more, even in the case that the stored episodes are optimal, it exist the risk that the agent just focuses on those episodes that might not well ensure the needed diversity as explained above (see an example in Figure \ref{fig:o2dlh_seeds_fixed_buffer}). Obviously, using the extrinsic reward for prioritization is just a criteria for classifying good episodes that is not mandatory, although has been broadly adopted as it is one the most important (if not the main) performance metric used to evaluate a RL algorithm.

% Close
%The aforementioned problems can persist even when using more levels.In that sense, it is difficult to ensure that the off-policy updates are going to be helpful although annealing its importance as long as the training progresses can be a good strategy (schedule the ratio dynamically) to minimize those drawbacks as we have empirically shown before.
The hyper-specialization of the buffer may turn the off-policy learning poor whatever the quantity of levels processed. In that sense, the off-policy updates could be dynamically scheduled to maximize the contribution along the training. Ideally, those updates should be selected from a buffer which fairly matches the optimal scores across levels.

\section{Conclusions}\label{sec:conc}

This work has postulated that the use of \textit{intrinsic motivation} can improve the sample efficiency of \textit{self-imitation learning} approaches in those sparse reward scenarios where the exploration needs hinder the collection of good episodes to be replayed. Based on this research hypothesis, a framework combining RAPID and BeBold has been proposed, exposing experimentally an equal or better sample efficiency when compared to RAPID, BeBold or SIL approaches on their own, when solving challenging tasks formulated in MiniGrid's procedurally-generated environments. We have shown that this advantage holds as long as the selected IM method is efficient enough. Furthermore, we highlight the necessity of scheduling correctly the on-policy and off-policy updates, as well as the use of a rollout size that considers multiple episodes in the same optimization step to reduce the variance and achieve an optimal policy. Finally, we have discussed the reasons why the off-policy updates do not necessarily have to be helpful due to 1) the environment requirements; 2) the diversity between the stored episodes; and 3) the way in which prioritization is applied, being subject to a reward function that does not distinguish between optimal solutions across levels due to the different steps required to solve them.
%optimal and suboptimal levels. 

In the future we aim to extend this study to continuous-state spaces, examining whether our insights herein offered can be extrapolated to other procedurally-generated benchmarks. Moreover, we plan to overcome the aforementioned limitations related to the off-policy updates, so that the agent's learning process can be accelerated without requiring a manual setup of the on-policy/off-policy schedule and the definition of the reward function itself.

\section*{Acknowledgments}

The authors would like to thank Daochen Zha from Rice University for the help provided in the implementation of RAPID. A. Andres receives funding support from the Basque Government through its BIKAINTEK PhD support program. J. Del Ser also acknowledges funding support from the Department of Education of the Basque Government (Consolidated Research Group MATHMODE, IT1456-22).

\bibliographystyle{IEEEtran}
\bibliography{IEEEexample}

\end{document}